%% file: CLEAR_JBHI.tex
\documentclass[journal]{IEEEtran}

\usepackage[nocompress]{cite}
\usepackage{graphicx}
\usepackage{amsmath}
\usepackage{amssymb}
\usepackage{array}
\usepackage{booktabs}
\usepackage{url}
\usepackage{xcolor}
\usepackage{titlesec}
\usepackage{needspace}
\usepackage[
  hidelinks,
  unicode=true,
  bookmarks=true,
  bookmarksnumbered=true,
  pdftitle={CLEAR: A Causal, Lightweight, and Explainable Artifact Removal Framework for Wearable ECG Monitoring},
  pdfauthor={Bowen Li, Yu Lu, Qingrui Deng, Liangyu Wu, Zhengyang Hao},
  pdfsubject={Manuscript for the IEEE Journal of Biomedical and Health Informatics},
  pdfkeywords={electrocardiography, artifact removal, wearable monitoring, quality assessment}
]{hyperref}
\usepackage{orcidlink}

\definecolor{ClearBlue}{RGB}{0,82,155}
\titleformat{\section}[block]
  {\normalfont\large\bfseries\centering\color{ClearBlue}}
  {\thesection.}{0.5em}{\MakeUppercase}
\titleformat{\subsection}[block]
  {\normalfont\normalsize\itshape\color{ClearBlue}}
  {\thesubsection}{0.5em}{}
\titleformat{\subsubsection}[block]
  {\normalfont\normalsize\itshape\color{ClearBlue}}
  {\thesubsubsection}{0.5em}{}

\begin{document}

\title{\color{ClearBlue}CLEAR: A Causal, Lightweight, and Explainable Artifact Removal Framework\\for Wearable ECG Monitoring}

\author{%
Bowen~Li\orcidlink{0009-0001-2277-8652},~\quad Yu~Lu\orcidlink{0000-0001-5208-7919},~\quad Qingrui~Deng\orcidlink{0009-0007-9214-5096},~\quad Liangyu~Wu\orcidlink{0009-0009-2264-9138},~and~Zhengyang~Hao\orcidlink{0009-0008-8164-0662}
\thanks{The authors are with Hangzhou Dianzi University, Hangzhou 310018, China. Corresponding author: Yu Lu (e-mail: y\_lu@hdu.edu.cn).}
\thanks{B. Li: libowen@hdu.edu.cn; Q. Deng: 2990979292@qq.com; L. Wu: 1979084276@qq.com; Z. Hao: haozhengyang@hdu.edu.cn.}
\thanks{Manuscript prepared for submission to the IEEE Journal of Biomedical and Health Informatics; received and revised dates to be completed by the journal.}}

\markboth{IEEE Journal of Biomedical and Health Informatics}%
{B. Li \MakeLowercase{\textit{et al.}}: CLEAR: A Causal, Lightweight, and Explainable Artifact Removal Framework for Wearable ECG Monitoring}

\maketitle

\begin{abstract}
Wearable electrocardiogram (ECG) monitoring requires artifact removal that preserves morphology, avoids future samples, and exposes model intervention. Most reconstruction-oriented denoisers address this tension primarily through reconstruction capacity: larger bidirectional or generative backbones reduce waveform error, while localization and quality, when predicted at all, are rarely produced by the same causal model. To address these issues, we propose CLEAR, a causal 28,330-parameter framework combining four dilated temporal convolutions with a unidirectional gated recurrent unit. It jointly estimates residual noise, pointwise masks, three-state quality, noise context, and global artifacts, supporting waveform recovery and quality-guided pass, denoise, or discard decisions. On PTB-XL clean waveforms contaminated with temporally separated MIT-BIH noise, the evaluation enforces patient-disjoint partitions, uses fixed test inputs across three seeds, and examines zero-fine-tuning transfer to three external databases. CLEAR achieved root-mean-square error of 0.3799 $\pm$ 0.0008, signal-to-noise-ratio improvement of 8.46 $\pm$ 0.03~dB, R-peak F1 of 0.9636 $\pm$ 0.0006, and pointwise mask F1 of 0.8573 $\pm$ 0.0056 on the fixed test, with external R-peak F1 ranging from 0.9564 to 0.9852 across three databases. Adapting only the 291-parameter quality head to human annotations raised quality macro F1 from 0.2444 to 0.8176 without changing denoising weights. A larger bidirectional comparator and a similarly compact U-Net achieved lower reconstruction error, while a power-line stress test exposed a failure mode. CLEAR provides an auditable balance among reconstruction, explanation, quality control, and resource constraints rather than claiming universally superior denoising accuracy.
\end{abstract}


\IEEEpeerreviewmaketitle

\input{sections/01_introduction}
\input{sections/02_related_work}
\input{sections/02_clear_framework}
\input{sections/03_experimental_protocol}
\input{sections/04_results}

\input{sections/05_ablation_discussion}
\input{sections/06_conclusion}

\bibliographystyle{IEEEtran}
\bibliography{references,references_additional}


\end{document}

%% file: sections/01_introduction.tex
\section{Introduction}\label{sec:introduction}

\IEEEPARstart{C}{ardiovascular} diseases caused an estimated 19.8 million deaths in 2022, approximately 32\% of global mortality~\cite{who2026cvd}. Wearable electrocardiogram (ECG) systems extend observation beyond brief clinical recordings, with compact recurrent models and armband systems enabling continuous or on-device analysis~\cite{saadatnejad2020lstm,lazaro2020armband}. Their value depends on reducing everyday interference without obscuring clinically relevant morphology. Because such interference is unavoidable in ambulatory recording, artifact handling is not an optional post-processing step but a precondition for using wearable ECG at all.

Prior work addresses the problem from three directions. Conventional filtering---band-pass, adaptive, wavelet, and Kalman variants---is transparent and inexpensive, but its fixed spectral assumptions break down when artifacts are transient or overlap physiologic components~\cite{chatterjee2020review,jia2024review,widrow1975adaptive,donoho1995denoising,huang1998emd,xu2019motion,nagai2017swt,smital2012waveletwiener,poungponsri2013adaptive,hesar2020kalman,sameni2007kalman,vanderbijl2022quality}. Deep denoisers, from convolutional and recurrent autoencoders to transformer and diffusion models, reach lower waveform error, but their computational assumptions and evaluation protocols differ widely, so reported scores are not interchangeable across studies~\cite{xiong2016stacked,chiang2019fcn,antczak2018recurrent,romero2021deepfilter,aghaomidi2022deeprtsnet,chen2024tcdae,li2024descod,ho2020ddpm,song2021score}. At the compact end, lightweight U-Net and recurrent designs target wearable budgets, and multitask formulations couple denoising with localization or quality assessment~\cite{ronneberger2015unet,qiu2022uldenet,sheu2022lightweight,hu2024lunet,dias2024cleaning,chen2023swindae,chen2024mtlnet,hu2025parallel}. What remains open is not reconstruction accuracy alone, but whether a compact model can expose its own processing order, its intervention region, and the trustworthiness of each segment under a leakage-controlled protocol. Section~\ref{sec:related} examines these directions in detail.

We introduce \textbf{CLEAR}, a causal, lightweight, and explainable artifact-removal framework containing 28,330 trainable parameters. Four dilated temporal convolutions and a unidirectional GRU predict residual contamination without future samples. Shared representations additionally yield pointwise masks, three-state quality, multilabel noise context, and global-artifact outputs. Quality-guided inference passes good segments, denoises noisy segments, or withholds severely degraded windows for manual review. All-noisy and all-denoised reference arms interpret retained performance together with coverage.

Evaluation combines PTB-XL clean waveforms and patient-aware benchmarking principles~\cite{wagner2020ptbxl,strodthoff2021benchmark} with temporally separated noise from the MIT-BIH Noise Stress Test Database~\cite{moody1984noise}. Seven MA/EM/BW combinations at four SNRs form an unchanged test set. Zero-fine-tuning transfer uses QTDB, MITDB, and ECG-ID~\cite{laguna1997qt,moody2001mitdb,lugovaya2005ecgid}. A separate experiment adapts only the quality head to human annotations from the Brno University of Technology ECG Quality Database (BUT QDB)~\cite{smital2026butqdb}. Four-View BiGRU and compact U-Net achieve stronger aggregate reconstruction, while a power-line test exposes a performance boundary; these outcomes delimit where CLEAR is and is not the right choice.

The contributions are:
\begin{enumerate}
  \item \textbf{We establish a leakage-controlled and auditable evaluation protocol.} Patient-disjoint sources, chronologically isolated noise, fixed test inputs, and external databases make each generalization claim explicit. The source split has zero cross-split patients; a record-level partition of the same cohort yields 165 overlapping patients across splits. Residuals, masks, context labels, requested/measured SNRs, and source metadata remain traceable.
  \item \textbf{We develop a compact causal multitask processor.} Its 28,330-parameter convolution--GRU backbone combines residual recovery with pointwise localization, quality, noise-context, and global-artifact prediction. Seven MA/EM/BW combinations and grouped augmentations support controlled training and ablation without attributing unsupported effects to individual perturbations.
  \item \textbf{We make quality-guided inference inspectable and independently adaptable.} Residuals, masks, context probabilities, and quality states explain where, what, and why the model processes. Pass/denoise/discard decisions include coverage and clean controls. Adapting only the 291-parameter quality head raises human-annotation macro F1 from 0.2444 to 0.8176 while freezing denoising weights.
  \item \textbf{We evaluate the complete operating profile and its boundaries.} Reconstruction, morphology, rhythm, auxiliary outputs, selective inference, ablations, efficiency, and cross-database transfer are evaluated jointly. BW-specific advantages are reported with stronger baselines, weak event localization, discarded-window costs, database-dependent rankings, and power-line performance limits.
\end{enumerate}

Section~\ref{sec:related} reviews related work, Section~\ref{sec:methods} defines CLEAR, Section~\ref{sec:experimental_protocol} describes the experimental protocol, Section~\ref{sec:results} reports comparative results, Section~\ref{sec:ablation_discussion} presents ablations, stress tests, and discussion, and Section~\ref{sec:conclusion} concludes the paper.

%% file: sections/02_related_work.tex
\section{Related Work}\label{sec:related}

Work on this problem falls into three groups: conventional filtering, learned waveform reconstruction, and joint processing that also predicts localization or quality. A fourth strand concerns how such systems are evaluated, since partitioning and test-input construction determine what a reported score means. The remainder of this section reviews each strand and states where CLEAR departs from it.

\subsection{From Conventional Filtering to Deep ECG Reconstruction}

ECG enhancement has long relied on band-pass filtering, adaptive filtering, and wavelet representations. These approaches remain useful baselines because their computational behavior and signal transformations are relatively transparent. Their limitations become important when artifacts are transient, nonstationary, or spectrally entangled with physiologic morphology~\cite{vanderbijl2022quality}. Fully convolutional denoising autoencoders demonstrated that hierarchical nonlinear representations can recover ECG waveforms under several controlled noise conditions~\cite{chiang2019fcn}. Subsequent architectures have pursued stronger reconstruction through two-stage robust networks, convolutional denoising autoencoders with transformer encoders, and score-based diffusion models~\cite{aghaomidi2022deeprtsnet,chen2024tcdae,li2024descod}.

These studies establish the breadth of the design space, from compact convolutional systems to iterative generative models. Their published scores are not directly interchangeable, however, when datasets, patient separation, noise realizations, target intensities, and evaluation units differ. We therefore do not import cross-paper numerical rankings into our primary comparison. Instead, CLEAR is evaluated against learning-based and conventional baselines under a shared data and test protocol, while the literature is used to position its functional and architectural choices.

\subsection{Lightweight and Online-Oriented ECG Processing}

Wearable applications impose restrictions on model storage, computation, and processing order. Work on wearable ECG classification has illustrated the practical value of small recurrent models~\cite{saadatnejad2020lstm}. For denoising, the BiGRU framework of Dias \textit{et al.} demonstrated that a compact recurrent architecture can handle diverse local and global artifact combinations and can be examined on independently acquired industrial recordings~\cite{dias2024cleaning}. Knowledge distillation has also been used to construct a lightweight U-Net for ECG denoising~\cite{qiu2022uldenet}, while LUNet explicitly combines a lightweight architecture with noise localization~\cite{hu2024lunet}.

CLEAR shares the objective of compact processing but makes a different architectural tradeoff. Causal dilated convolutions and unidirectional recurrence avoid future-sample dependence, and residual prediction exposes the estimated contamination directly. The additional heads increase the parameter count slightly relative to the 26,121-parameter Dias architecture, rather than establishing a new parameter-reduction record. Moreover, parameter count is treated as a design constraint, not as a substitute for measurements of wearable-device latency, memory, or energy consumption.

\subsection{Joint Denoising, Localization, and Quality Assessment}

A waveform-only denoiser leaves open whether a segment should be trusted, retained, or routed for further review. SwinDAE uses a one-dimensional Swin transformer and denoising autoencoder for quality-related representation learning~\cite{chen2023swindae}. MTL-NET explicitly exploits the relationship between signal quality assessment and denoising through multitask learning~\cite{chen2024mtlnet}. More recently, Hu \textit{et al.} presented a parallel framework for fine-grained localization, denoising, and signal quality assessment~\cite{hu2025parallel}. These methods demonstrate that joint learning is itself an established and active direction.

The distinction of CLEAR is the combination and auditing of several elements within a compact causal model: pointwise local-artifact masks, noise-type and global-artifact context, residual noise, and quality-driven pass/denoise/discard decisions. Its evaluation separates pointwise from event-level localization and distinguishes performance on all samples from performance on an adaptively retained subset. Human-annotated quality adaptation is reported separately from zero-fine-tuning denoising, preventing two different generalization claims from being conflated.

\subsection{Evaluation Credibility and Failure Visibility}

Patient-aware benchmarks such as PTB-XL emphasize that model performance must be interpreted together with data partitioning and evaluation design~\cite{strodthoff2021benchmark}. Synthetic denoising introduces two additional protocol questions: whether the noise recordings are temporally separated between training and evaluation, and whether all compared methods receive identical test perturbations. External databases, clean-signal controls, and downstream physiological measurements provide complementary checks, but each addresses a different failure mode rather than guaranteeing clinical validity.

Selective classification more generally formalizes the relationship between prediction coverage and risk~\cite{geifman2017selective}. CLEAR borrows the operational distinction between retained and rejected inputs, but does not assume a theoretical risk--coverage guarantee. Its quality-guided policy must therefore be compared with all-noisy and all-denoised processing, with coverage and discard rates reported alongside retained-sample performance.

CLEAR accordingly treats evaluation as part of the method. Patient and noise isolation define the target of generalization; fixed test inputs define the comparison; and external evaluation tests transfer beyond the training source. Negative results define where the method should not be applied. This perspective motivates reporting a stronger baseline's reconstruction advantage, the discard cost of selective inference, weak event-level localization, and power-line degradation alongside the principal positive findings.

%% file: sections/02_clear_framework.tex
\section{CLEAR Framework and Artifact-Removal Method}\label{sec:methods}

\begin{figure*}[!t]
  \centering
  \includegraphics[width=0.73\textwidth]{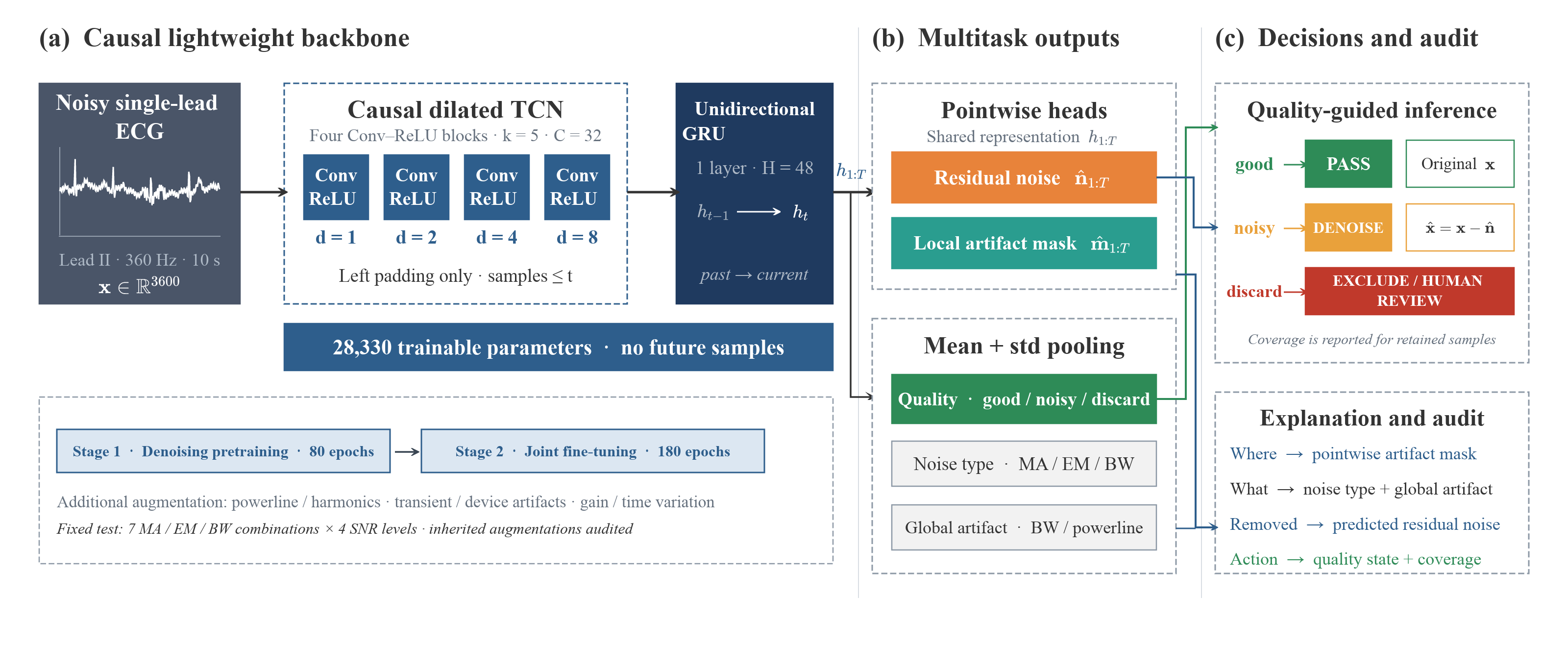}
  \caption{CLEAR architecture and inference. A causal dilated temporal convolutional network (TCN) and unidirectional gated recurrent unit (GRU) estimate residuals, pointwise masks, quality, noise context, and global artifacts. Quality states guide pass/denoise/discard decisions; residuals, masks, context, and coverage support auditing. The lower-left panel summarizes optimization, augmentation, fixed-test evaluation, and quality-head adaptation on the Brno University of Technology ECG Quality Database (BUT QDB).}
  \label{fig:clear_framework}
\end{figure*}

\subsection{Problem Formulation and Output Semantics}

\begin{figure}[!t]
  \centering
  \includegraphics[width=0.90\columnwidth]{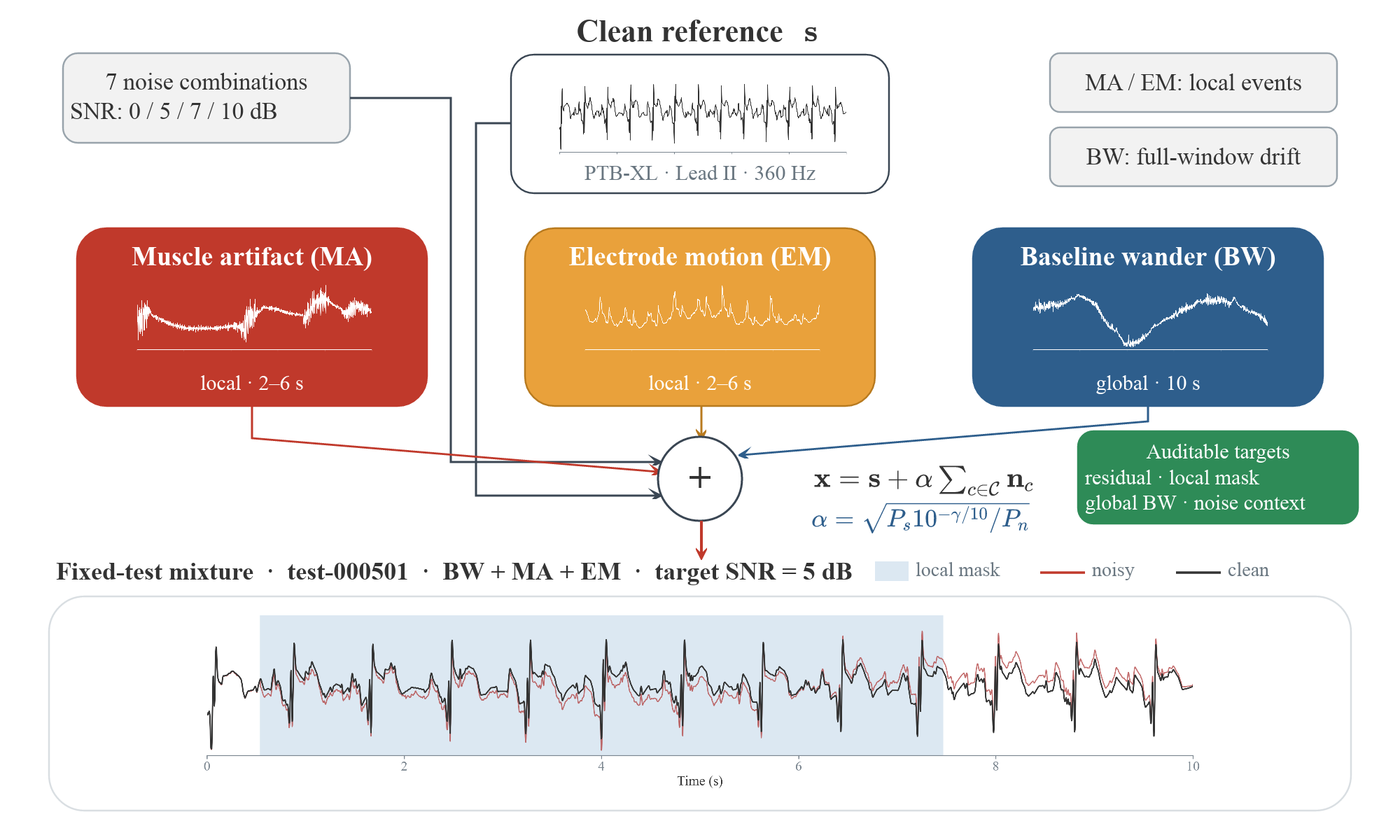}
  \caption{Local/global contamination and residual learning. Measured MA and EM occupy local intervals, whereas BW affects the full window. SNR-calibrated combinations provide paired noisy/clean waveforms with residual, mask, context, and global-artifact targets. The lower trace is a fixed-test BW+MA+EM example at requested 5-dB SNR; shading denotes the local-artifact mask.}
  \label{fig:noise_principle}
\end{figure}

Given a noisy single-lead ECG window $\mathbf{x}\in\mathbb{R}^{L}$, CLEAR estimates its clean reference $\mathbf{s}\in\mathbb{R}^{L}$ together with auxiliary outputs describing contamination and usability. Main experiments use $L=3600$, a 10-s interval at 360~Hz. With $\mathbf{n}=\mathbf{x}-\mathbf{s}$, the network predicts contamination rather than directly generating the cleaned waveform:
\begin{equation}
  \widehat{\mathbf{n}}=f_{\theta}(\mathbf{x}),\qquad
  \widehat{\mathbf{s}}=\mathbf{x}-\widehat{\mathbf{n}}.
  \label{eq:residual_recovery}
\end{equation}
Both the recovered waveform and estimated intervention therefore remain available for inspection. Fig.~\ref{fig:noise_principle} illustrates local MA/EM events and global BW with their paired supervision targets.

A pointwise mask $\widehat{\mathbf{m}}\in[0,1]^{L}$ identifies local MA/EM samples. Window-level outputs comprise three quality logits, three independent MA/EM/BW context logits, and two global BW/power-line logits. These window-level outputs prevent global contamination from being forced into an artificially sparse interval mask. Main quality targets are deterministic simulation labels; adaptation to human annotations is evaluated separately.

\noindent\textbf{What causal, lightweight, and explainable mean here.} The three adjectives in the title are used in checkable senses. \emph{Causal} means that every sample-level output depends only on current and past samples (Section~\ref{sec:methods}, Eq.~\eqref{eq:causal_convolution}), while quality, context, and global decisions are deliberately segment-level; the resulting streaming behaviour is verified in Section~\ref{sec:stress_results}. \emph{Lightweight} means a fixed budget, not a comparison: 28,330 parameters, 116.54~KiB in FP32, and 61.12~ms per window on one CPU core, with no claim about microcontroller feasibility. \emph{Explainable} means that each window exposes four independently checkable quantities---the removed residual, the intervention interval, the assigned noise type, and the quality state that selects pass, denoise, or discard---so an explanation is these outputs and their consistency, not a post-hoc saliency map; Section~\ref{sec:auxiliary_results} tests the consistency by measuring removed energy outside the predicted mask.

\subsection{Causal Lightweight Backbone}

The architecture follows from four constraints rather than from capacity. Ambulatory monitoring forbids future-sample dependence, so every operator is either locally causal or recurrent in the forward direction; a bidirectional or attention-based alternative would violate this constraint regardless of accuracy. Compact devices bound storage and energy, so the temporal model is built from depthwise-cheap dilated convolutions with a narrow channel count and a single recurrent layer, and the input is processed as a raw single-lead window without hand-crafted spectral features. Because contamination differs in scale and extent, local MA/EM events and whole-window BW require different receptive fields; stacked dilations of 1, 2, 4, and 8 provide this range with only four layers. Finally, a residual parameterization makes the estimated intervention directly inspectable: the network predicts the noise to be subtracted, so the mask, the recovered waveform, and the removed component all describe the same quantity. These four requirements jointly determine the backbone, the head structure, and the choice of a shared representation for the waveform and window-level tasks, which we describe next.

Fig.~\ref{fig:clear_framework} summarizes the information flow. Four temporal convolutional layers extract local morphology, followed by a unidirectional GRU that propagates information forward. The design follows temporal convolutional and GRU principles~\cite{bai2018tcn,cho2014gru}. Each convolution has kernel size five, 32 output channels, ReLU activation, and dilation $d_l\in\{1,2,4,8\}$. Left-only zero padding gives
\begin{equation}
  h_t^{(l)}=\operatorname{ReLU}\!\left(
  \sum_{k=0}^{4}W_k^{(l)}h_{t-kd_l}^{(l-1)}+b^{(l)}
  \right),
  \label{eq:causal_convolution}
\end{equation}
where positions before the sequence are zero. Every output at $t$ thus depends only on current and past samples. A single-layer unidirectional GRU with hidden width 48 produces $\mathbf{h}_{1:T}$ without backward recurrence.

Pointwise linear heads produce residual noise and a local-mask logit, using a 0.5 sigmoid threshold for binary masks. For window-level inference, temporal mean and standard deviation are concatenated:
\begin{equation}
  \mathbf{z}=\left[\operatorname{mean}_{t}(\mathbf{h}_{t});
  \operatorname{std}_{t}(\mathbf{h}_{t})\right].
  \label{eq:temporal_pooling}
\end{equation}
A shared linear readout produces quality, context, and global-artifact logits. Softmax is used for quality and independent sigmoids for multilabel outputs. The complete model has 28,330 trainable parameters.

\subsection{Multitask Learning Objective}\label{sec:objective}

We denote the resulting training criterion the CLEAR objective; it couples recovery with localization, quality, context, and explanation consistency:
\begin{equation}
\begin{aligned}
  \mathcal{L}={}&\mathcal{L}_{\mathrm{den}}
  +0.35\mathcal{L}_{\mathrm{mask}}
  +0.20\mathcal{L}_{\mathrm{qc}}
  +0.25\mathcal{L}_{\mathrm{corr}}\\
  &+0.02\mathcal{L}_{\mathrm{smooth}}
  +0.20\mathcal{L}_{\mathrm{global}}
  +0.20\mathcal{L}_{\mathrm{outside}} .
  \label{eq:joint_objective}
\end{aligned}
\end{equation}
To emphasize higher-amplitude physiologic regions without manual QRS intervals, the clean reference defines
\begin{equation}
  w_t=1+2\,\mathbb{I}\!\left(|s_t|>
  \operatorname{median}_{\tau}|s_{\tau}|\right),
  \label{eq:sample_weight}
\end{equation}
The median threshold is a mild emphasis on higher-amplitude samples rather than a QRS detector: roughly half of the samples receive weight $3$, so the term raises the cost of errors in large-amplitude regions without isolating QRS complexes. A detector-driven or high-percentile ($90$--$95$th) weighting would concentrate the weight more sharply; because the weighting interacts with the two-stage schedule, we retain the median form for consistency with the reported runs and report the alternative as an open design choice.
and the weighted residual objective is
\begin{equation}
  \mathcal{L}_{\mathrm{den}}=
  \frac{\sum_t w_t\left(\widehat{n}_t-x_t+s_t\right)^2}
  {\sum_t w_t+\epsilon}.
  \label{eq:denoising_loss}
\end{equation}
Here $\mathcal{L}_{\mathrm{mask}}$ is binary cross-entropy with logits, $\mathcal{L}_{\mathrm{qc}}$ combines quality cross-entropy with multilabel context loss, and $\mathcal{L}_{\mathrm{global}}$ applies binary cross-entropy to global BW and power-line outputs.

Correlation and second-difference terms preserve waveform shape:
\begin{equation}
  \mathcal{L}_{\mathrm{corr}}=
  1-\operatorname{corr}(\widehat{\mathbf{s}},\mathbf{s}),
  \label{eq:correlation_loss}
\end{equation}
\begin{equation}
  \mathcal{L}_{\mathrm{smooth}}=\frac{1}{L-2}
  \sum_{t=2}^{L-1}
  \left(\widehat{s}_{t-1}-2\widehat{s}_t+
  \widehat{s}_{t+1}\right)^2 .
  \label{eq:smoothness_loss}
\end{equation}
Finally, $\mathcal{L}_{\mathrm{outside}}$ penalizes residual predictions outside the true local mask for windows without a global artifact, aligning the intervention with local contamination while permitting nonlocal correction when a global artifact is present:
\begin{equation}
  \mathcal{L}_{\mathrm{outside}}=
  \frac{\sum_t \widehat{n}_t^{2}\,(1-m_t)\,\mathbb{I}[\text{no global artifact}]}
       {\sum_t (1-m_t)\,\mathbb{I}[\text{no global artifact}]+\epsilon},
  \label{eq:outside_loss}
\end{equation}
where $m_t$ is the local-artifact target and $\epsilon=10^{-8}$ guards the all-masked case.

\subsection{Two-Stage Optimization}

Denoising pretraining runs for at most 80 epochs with residual, correlation, and outside-residual terms; mask, quality/context, global, and smoothness weights are zero. It uses learning rate $10^{-3}$ and exponential moving-average decay 0.995. Joint fine-tuning runs for at most 180 epochs with the complete objective, learning rate $3\times10^{-4}$, and decay 0.999. Both stages use AdamW~\cite{loshchilov2019adamw}, weight decay 0.01, batch size 256, five warmup epochs, gradient clipping at 1.0, and early-stopping patience 15 and 30 epochs, respectively. Checkpoints use validation data only; three seeds quantify run-to-run variability. Superiority over randomly initialized joint training is not assumed; Section~\ref{sec:single_stage} reports a direct comparison at the same epoch budget.

\subsection{Quality-Guided Selective Inference}

The quality state determines the returned output:
\begin{equation}
  \mathcal{D}(\mathbf{x})=
  \begin{cases}
    \mathbf{x}, & q=\mathrm{good}\quad(\mathrm{pass}),\\
    \widehat{\mathbf{s}}, & q=\mathrm{noisy}\quad(\mathrm{denoise}),\\
    \mathrm{not\ retained}, & q=\mathrm{discard}.
  \end{cases}
  \label{eq:quality_decision}
\end{equation}
Discard means that the window is withheld from automatic output and routed for manual review; it does not mean permanent deletion of the original recording. This policy is related to selective classification and coverage--risk analysis~\cite{geifman2017selective}, but assumes no theoretical risk--coverage guarantee. Adaptive inference is compared with returning every noisy input and denoising every input. Coverage is $N_{\mathrm{retained}}/N$, and retained-subset metrics are never presented as full-cohort performance.

%% file: sections/03_experimental_protocol.tex
\section{Data and Experimental Protocol}\label{sec:experimental_protocol}

\subsection{Datasets and Signal Preparation}\label{sec:datasets}

The protocol separates three sources of generalization: unseen patients, unused noise intervals, and unseen ECG databases. Clean records came from PTB-XL~\cite{wagner2020ptbxl} through PhysioNet~\cite{goldberger2000physionet}: 21,799 10-s Lead-II records from 18,869 patients. Signals were resampled from 500 to 360~Hz by polyphase anti-alias filtering and filtered offline with a second-order zero-phase Butterworth bandpass (1--45~Hz) using SciPy~\cite{virtanen2020scipy}. This reference-preparation filter is offline and forms no part of causal inference. Each record was median-centered, divided by $1.4826\times\mathrm{MAD}$, and clipped at $\pm5$.

Two qualifications follow. First, the 1--45~Hz band is a controlled-benchmark reference, not a diagnostic acquisition band: guidelines recommend low-frequency response near 0.05~Hz and high-frequency response near 150~Hz~\cite{kligfield2007aha}, so all morphology-preservation statements are \emph{relative to the 1--45~Hz processed reference} rather than to the raw acquired signal. Second, the scaling statistics are computed from the clean source and are unavailable to a device observing only the contaminated input; the reported results therefore use an oracle-scaled representation, and a deployable variant must substitute fixed training-set statistics or an online causal estimator.

To test whether the oracle statistics are load-bearing, a deployable variant was trained and evaluated end to end. A single pair of statistics was pooled over the training split only ($\mathrm{median}=-0.0186$, $\mathrm{MAD}=0.0359$) and substituted for the per-record values at scaling time; no other component changed. The fixed-statistics model reached RMSE $0.3711$ versus $0.3799$ for the oracle-scaled model ($-2.3\%$), $\Delta$SNR $8.358$ versus $8.461$~dB ($-1.2\%$), R-peak F1 at 50~ms tolerance $0.9443$ versus $0.9333$ ($+1.2\%$), and mask F1 $0.8592$ versus $0.8573$ ($+0.2\%$) on the same fixed test, with per-record scale factors of $0.63$ versus $0.67$ in MAD units and a clipping rate rising from $4.13\%$ to $5.34\%$. The four changes are small in absolute terms and mixed in direction, three improving and one degrading. The variant was trained once, so its $0.0088$ RMSE change is about eleven times the three-seed standard deviation of the oracle model; the comparison is therefore reported as a direction, not as statistical equivalence.

Both channels of the MIT-BIH BW, EM, and MA recordings~\cite{moody1984noise} were divided chronologically into nonoverlapping 80\%/10\%/10\% portions, containing 520,000, 65,000, and 65,000 samples per channel at 360~Hz. Patient assignment and noise time spans therefore remain disjoint across corresponding partitions.

Four independently seeded contamination views were generated for each training and validation record at requested base SNRs of 0, 5, 7, and 10~dB. Together with the unchanged fixed test, this produced 77,248 samples: 60,636 training, 13,296 validation, and 3,316 test windows. The same test waveforms and masks were preserved byte-for-byte for every method and seed. The contamination bank includes MA/EM/BW combinations, power-line interference and harmonics, transient bursts, electrode pops, saturation, packet loss, and time-varying intensity. These augmentations affect four-view generation and a subset of the fixed test; 524 test windows (15.8\%) contain power-line interference. Requested and measured SNRs were stored separately.

Post-training transfer used QTDB, MITDB, and ECG-ID~\cite{laguna1997qt,moody2001mitdb,lugovaya2005ecgid}, comprising 1,080, 576, and 1,240 controlled test views after excluding 15 duplicate-source QT/MIT views. No denoising weight was fine-tuned. The Brno University of Technology ECG Quality Database (BUT QDB)~\cite{smital2026butqdb} was reserved for human-quality adaptation. Only the quality classifier was trained; the backbone and all denoising, localization, context, and global-artifact weights remained frozen. BUT QDB is therefore excluded from zero-fine-tuning denoising transfer.

\subsection{Contamination Protocol and Data Partition}\label{sec:contamination_protocol}

For clean record $\mathbf{s}$, the observation is
\begin{equation}
  \mathbf{x}=\mathbf{s}+\alpha\sum_{c\in\mathcal{C}}\mathbf{n}_c
  +\mathbf{a}_{\mathrm{aug}},
  \label{eq:contamination_model}
\end{equation}
where $\mathcal{C}$ covers MA, EM, BW, MA--EM, MA--BW, EM--BW, and MA--EM--BW, and $\mathbf{a}_{\mathrm{aug}}$ denotes additional augmentations. MA and EM occupy local intervals lasting 2--6~s; BW affects the full window. Augmentations add 50/60-Hz interference with up to three harmonics, bursts lasting 0.05--0.5~s, electrode pops, saturation, packet loss lasting 0.05--0.3~s, and time-varying MA/EM intensity. Five percent of nonsevere training and validation views remain clean pass-through examples.

Before augmentation, the measured noise mixture is scaled to requested base SNR $\gamma$:
\begin{equation}
  \alpha=\sqrt{\frac{P_s\,10^{-\gamma/10}}{P_n}},\qquad
  P_s=\frac{1}{L}\sum_t s_t^2,\quad
  P_n=\frac{1}{L}\sum_t n_t^2 .
  \label{eq:snr_scaling}
\end{equation}
Four views use $\gamma\in\{0,5,7,10\}$~dB, and final SNR is recomputed after all events. Audit targets include residual, local mask, MA/EM/BW context, and global BW/power-line labels. Deterministic rules assign \emph{discard} to saturation, packet loss, or measured SNR below 0~dB; \emph{good} to clean windows or measured SNR of at least 10~dB; and \emph{noisy} otherwise. These labels are not equated with human judgments.

Table~\ref{tab:data_partition} summarizes the organization. Patient assignment precedes view generation and yields zero cross-split patients; a record-level partition of the identical cohort yields 165 overlapping patients across splits. Fixed-test waveforms and combined masks are reused without regeneration, while finer labels are accepted only when deterministic replay reproduces both exactly. The stricter partition improves evaluation credibility, not denoising accuracy by itself.

\begin{table}[!t]
  \centering
  \caption{Patient- and Noise-Isolated Data Organization. Parentheses Give Within-Split Percentages.}
  \label{tab:data_partition}
  \scriptsize
  \setlength{\tabcolsep}{2.6pt}
  \renewcommand{\arraystretch}{1.08}
  \begin{tabular}{@{}lrrr@{}}
    \toprule
    & Training & Validation & Fixed test \\
    \midrule
    Patients & 13,145 & 2,853 & 2,871 \\
    Source records & 15,159 & 3,324 & 3,316 \\
    Views per source & 4 & 4 & 1 (fixed) \\
    Target SNR (dB) & 0/5/7/10 & 0/5/7/10 & 0/5/7/10 \\
    Final samples & 60,636 & 13,296 & 3,316 \\
    Noise time share & 80\% & 10\% & 10\% \\
    Noise samples/channel & 520,000 & 65,000 & 65,000 \\
    Clean-preserved views & 3,076 & 635 & 0 \\
    Power-line views & 8,657 (14.3\%) & 1,891 (14.2\%) & 524 (15.8\%) \\
    \bottomrule
  \end{tabular}
\end{table}

\subsection{Comparison Methods and Implementation Details}\label{sec:comparison_methods}

All learned models use identical training/validation views and fixed test inputs. CLEAR uses seeds 0, 1, and 2; Four-View BiGRU and lightweight 1D U-Net use common seeds 0 and 1, and their two-seed variability is not merged with CLEAR's three-seed summary. Model selection uses validation data only.

Four-View BiGRU has two bidirectional GRU layers of hidden width 64 and a linear readout, totaling 100,353 parameters. It uses batchwise RMSE, Adam~\cite{kingma2015adam}, learning rate $5\times10^{-3}$, at most 130 epochs, and patience 20. It tests reconstruction capacity with future context rather than causal deployability. The independent 1D U-Net~\cite{ronneberger2015unet} contains 29,928 parameters, residual convolutional blocks with GroupNorm and SiLU, two downsampling/upsampling stages, and skip connections with widths 16, 24, and 31. It uses MSE, Adam, learning rate $10^{-3}$, at most 150 epochs, and patience 25, providing a closely matched nonrecurrent budget.

CLEAR follows Section~\ref{sec:methods}; all learned methods use batch size 256. Baseline checkpoints use validation RMSE, whereas CLEAR uses its validation multitask composite. Denoise-only and auxiliary-removal variants examine the effect of this difference.

Wavelet and bandpass references are selected by validation RMSE and frozen before testing. Wavelet candidates combine Symlet-8 or Daubechies-6 wavelets, six levels, and universal or Bayes soft thresholding using PyWavelets~\cite{lee2019pywavelets}. Fourth-order bandpass candidates use low cutoffs 0.05 or 0.5~Hz and high cutoffs 40 or 100~Hz. Implementations use Python 3.11 with PyTorch 2.15.0, NumPy 2.4.6, SciPy 1.17.1, pandas 3.1.1, scikit-learn 1.9.1, PyWavelets 1.9.0, NeuroKit2 0.3.0, wfdb 5.0.0, ONNX 1.23.2 and ONNX Runtime 1.32.0~\cite{paszke2019pytorch}; latency is reported only with its benchmark context. All classical references (wavelet and band-pass) use zero-phase filtering (\texttt{filtfilt}) and are therefore non-causal comparators rather than causal deployable methods; a forward-only IIR variant with matched magnitude response is left to future work.

\subsection{Evaluation Metrics}\label{sec:evaluation_metrics}

Unless stated otherwise, waveform results use the all-denoised arm on all 3,316 fixed-test samples. Metrics are computed per sample, averaged within each dataset or noise--SNR stratum, and summarized across seeds by mean and sample standard deviation. External databases remain separate. Define $e_t=\widehat{s}_t-s_t$ for clean $\mathbf{s}$; $\mathbf{x}$ is the observed input:
\begin{equation}
  \mathrm{RMSE}=\sqrt{L^{-1}\sum_t e_t^2},\qquad
  \mathrm{PRD}=100\sqrt{\frac{\sum_t e_t^2}{\sum_t s_t^2+\epsilon}},
  \label{eq:rmse_prd}
\end{equation}
\begin{equation}
  \Delta\mathrm{SNR}=10\log_{10}\!
  \left(\frac{\sum_t(x_t-s_t)^2+\epsilon}
  {\sum_t(\widehat{s}_t-s_t)^2+\epsilon}\right).
  \label{eq:snr_improvement}
\end{equation}
The signed SNR definition preserves negative outcomes.

Morphology evaluation includes QRS-proxy absolute slope error adjacent to the mask in~\eqref{eq:sample_weight}, derivative correlation, 256-sample log-magnitude STFT distance, and matched R-peak timing error. PTB-XL analyses are stratified by NORM, MI, STTC, CD, and HYP when available. These are consistency measures, not expert PQRST, QT, or ST-interval measurements.

Pointwise mask precision, recall, and F1 use threshold 0.5, with
\begin{equation}
  P=\frac{\mathrm{TP}}{\mathrm{TP}+\mathrm{FP}},\quad
  R=\frac{\mathrm{TP}}{\mathrm{TP}+\mathrm{FN}},\quad
  F_1=\frac{2PR}{P+R},
  \label{eq:f1}
\end{equation}
where a sample counts as true positive when the prediction and the target agree at that sample. Classifier quality is summarized by the area under the receiver-operating characteristic curve (AUROC) computed from score rankings, and calibrated reliability by the expected calibration error
\begin{equation}
  \mathrm{ECE}=\sum_{b=1}^{B}\frac{|S_b|}{N}\left|\mathrm{acc}(S_b)-\mathrm{conf}(S_b)\right|,
  \label{eq:ece}
\end{equation}
over $B=10$ equal-width confidence bins, where $\mathrm{conf}(S_b)$ is the mean predicted probability in bin $S_b$. Event evaluation removes runs shorter than 10 samples and matches events one-to-one only when onset and offset are both within 100~ms. Quality uses three-class macro-F1 and 15-bin expected calibration error (ECE)~\cite{guo2017calibration}; context and global outputs use AUROC~\cite{fawcett2006roc} and, where applicable, F1. Undefined AUROC is excluded. Human-quality adaptation also reports subject-macro F1 and Cohen's $\kappa$~\cite{cohen1960kappa}.

NeuroKit2 detects R peaks in clean, noisy, and recovered waveforms~\cite{makowski2021neurokit2}. Primary one-to-one tolerance is 150~ms; F1, missed/wrong-peak rates, and median absolute RR error accompany SDNN, RMSSD, and LF/HF absolute differences. Ten-second HRV measures are within-record consistency quantities, not stable clinical estimates. Sensitivity analyses use 50/100/150-ms tolerances and Pan--Tompkins detection~\cite{pan1985qrs,hamilton1986qrs}. References are detector outputs on clean waveforms, not manual annotations.

Selective inference compares all-noisy, all-denoised, and quality-guided arms using
\begin{equation}
  \mathrm{coverage}=\frac{N_{\mathrm{retained}}}{N},\qquad
  \mathrm{discard\ rate}=1-\mathrm{coverage}.
  \label{eq:coverage}
\end{equation}
Adaptive metrics apply only to retained windows, with noisy and all-denoised references restricted to the same subset. Patient-macro averages and 1,000 patient-level bootstrap repetitions estimate the adaptive-minus-noisy RMSE difference and 95\% interval. Clean controls report alteration RMSE and pass rate.

Efficiency uses trainable parameters, serialized size, and single-window CPU latency with input $1\times3600\times1$, 10 warm-ups, and 50 timed repetitions. Dynamic INT8 quantization of linear and GRU layers is evaluated separately from FP32. These measurements do not establish wearable-device energy consumption.

\subsection{Ablation and Stress-Test Protocols}\label{sec:ablation_stress}

Eight variants isolate supervision, architecture, and data generation: denoise-only; without mask supervision; without quality/context/global supervision; TCN-only (16,234 parameters); GRU-only (8,218); without noise augmentation; without correlation/smoothness regularization; and without outside-mask explanation consistency. All retain patient/noise partitions, the fixed test, equivalent pretraining, and seeds 0--2. Removing a supervision term also removes its checkpoint-selection term. Paired same-seed differences prevent independently permuted comparisons; unavailable untrained outputs are marked as such rather than scored.

Stress interventions are applied independently without parameter updates: power-line interference, burst, electrode pop, saturation, packet loss, and a smooth time-varying envelope on the original residual noise. For stressor $k$,
\begin{equation}
  D_k=\frac{\mathrm{RMSE}(\widehat{\mathbf{s}}^{(k)},\mathbf{s})}
  {\max\!\left(\mathrm{RMSE}(\widehat{\mathbf{s}},\mathbf{s}),\epsilon\right)} .
  \label{eq:stress_degradation}
\end{equation}
This ratio compares stressed output with the model's own base output, not improvement over the stressed noisy input. The dedicated power-line analysis therefore additionally reports $\Delta\mathrm{SNR}$ and R-peak changes relative to its contaminated input. Fundamental and first three harmonic powers are measured on input/output residuals using Hann periodograms integrated over $\pm0.5$~Hz. Negative responses remain in the evaluation and are not described as cross-database transfer.

%% file: sections/04_results.tex
\section{Results}\label{sec:results}

\input{tables/main_morphology}
\input{tables/main_selective}
\input{tables/efficiency}
\input{tables/ablation_main}
\input{tables/external_common_seed}

\input{tables/rpeak_sensitivity}
\input{tables/main_auxiliary}

Unless explicitly identified as retained-subset analysis, waveform metrics include every sample in the corresponding fixed test. Comparator results use common seeds 0 and 1; CLEAR stability, morphology, auxiliary outputs, and ablations additionally include seed 2. Sample standard deviations describe run-to-run variation on unchanged inputs, not confidence intervals over patients.

\subsection{Training Convergence and Comparator Behavior}\label{sec:training_results}

The recorded, unsmoothed histories are summarized here rather than plotted; the per-seed curves are provided as Fig.~S1 in the supplementary material. CLEAR completed 80 pretraining epochs per seed (selecting epochs 79, 80, and 80) and 180 joint epochs (selecting 178, 180, and 171). Test RMSE was 0.3797, 0.3794, and 0.3808; the three-seed standard deviation was 0.20\% of the mean, and the three seed histories nearly overlap. Four-View BiGRU stopped after 84 and 77 epochs, selecting 64 and 57, with test RMSE 0.1817 and 0.1909. U-Net ran 150 epochs per seed, selecting 150 and 149, with RMSE 0.3623 and 0.3614. Validation RMSE was used within each denoising run; historical single-view and four-view validation values are not treated as architecture-only rankings because their validation views differ.

The deterministic references were also frozen by validation performance. Daubechies-6 wavelets with six levels and universal soft thresholding gave the best wavelet validation RMSE (0.9941); the fourth-order 0.5--40-Hz bandpass gave 0.6884. These configurations were applied unchanged to every test database, so their weak results cannot be attributed to test-set selection.

\subsection{Main Denoising Performance}\label{sec:main_results}

Table~\ref{tab:main_denoising} reports all-denoised performance on all 3,316 windows; its CLEAR row is the common-seed value so that every learned method is compared on the same two seeds, whereas the three-seed statistics quoted below describe run-to-run stability. CLEAR's three-seed RMSE was $0.3799\pm0.0008$, $\Delta$SNR $8.46\pm0.03$~dB, PRD $22.58\pm0.05$\%, and R-peak F1 $0.9636\pm0.0006$. Relative to the noisy input, CLEAR reduced RMSE by 63.92\%. Under common seeds, CLEAR reduced RMSE by 63.83\% relative to wavelet and 46.09\% relative to bandpass; wavelet slightly reduced R-peak F1, whereas bandpass and CLEAR improved it. Supplementary shape-and-error descriptors on the same fixed test gave SSD $593.18$, MAD $0.2140$, PRD $22.55\%$, and cosine similarity $0.9711$; the PRD value agrees with the seed-averaged entry in Table~\ref{tab:main_denoising} to within $0.03$ percentage points.

\input{tables/main_denoising}

The official-structure GRU control (26,121 parameters) produced RMSE $1.1752\pm0.0010$, $\Delta$SNR $-1.7057\pm0.0084$~dB, and R-peak F1 $0.7472\pm0.0084$ (Table~\ref{tab:main_denoising}). CLEAR reduced RMSE by 67.71\% relative to this tested reimplementation, without implying that the architecture would fail under every optimization protocol.

Patient-clustered paired bootstrap intervals were computed for the two headline comparisons (2{,}000 resamples over 2{,}871 patients, seed 0). CLEAR was worse than Four-View BiGRU by $\Delta$RMSE $0.198$ with a 95\% interval of $[0.195,0.201]$ and by $\Delta$SNR $6.54$~dB; it was worse than the 1D U-Net by $\Delta$RMSE $0.0174$ with an interval of $[0.0147,0.0200]$ and by $\Delta$SNR $0.32$~dB. All intervals exclude zero, so the reconstruction ranking is not an artefact of patient composition.

The 100,353-parameter Four-View BiGRU achieved the strongest reconstruction: RMSE $0.1863\pm0.0065$, $\Delta$SNR $14.8006\pm0.2954$~dB, and R-peak F1 $0.9788\pm0.0004$. Its network was unchanged from the historical Single-View BiGRU; four-view contamination training reduced mean RMSE from 0.3773 to 0.1863 (50.62\%) and increased $\Delta$SNR by 5.0990~dB. This is a training-protocol improvement to the earlier baseline, not a new CLEAR backbone. Relative to CLEAR, it uses 3.542 times as many parameters, future context, and a reconstruction-only objective. Its advantage is compatible with all three differences and cannot be attributed to parameter count alone.

The 29,928-parameter U-Net achieved RMSE $0.3619\pm0.0006$, $\Delta$SNR $8.8089\pm0.0180$~dB, and R-peak F1 $0.9689\pm0.0001$, exceeding CLEAR's aggregate reconstruction at a similar budget. Thus, a smaller parameter count is not by itself an accuracy advantage; CLEAR's intended distinction is causal inference with residual, mask, context, and quality outputs in one model.

\begin{figure*}[!t]
  \centering
  \includegraphics[width=0.73\textwidth]{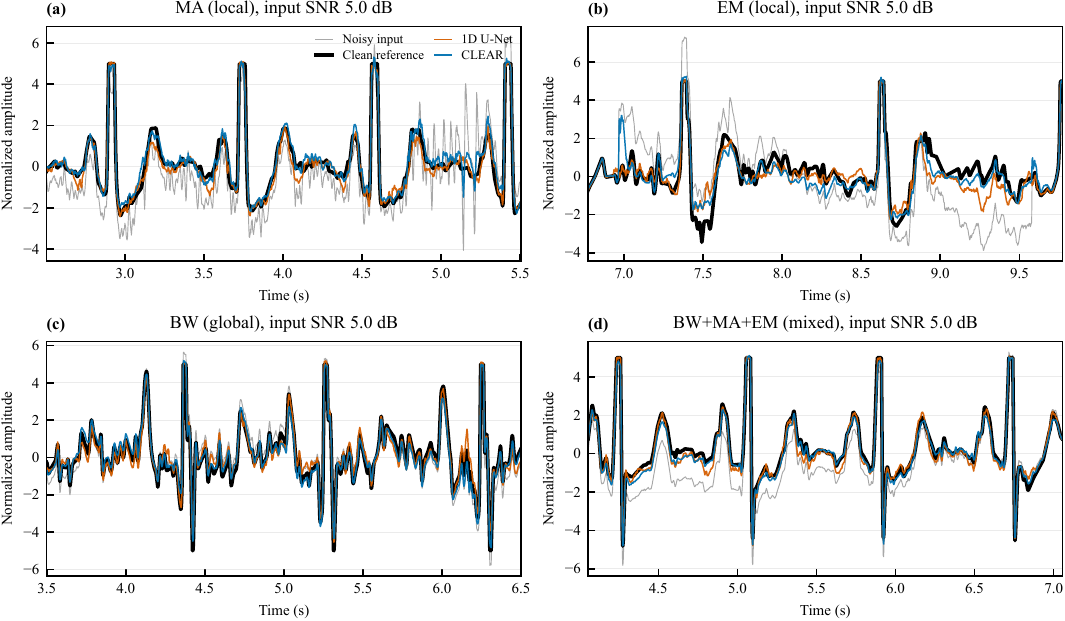}
  \caption{Qualitative denoising comparison on fixed-test windows at approximately 5-dB input SNR (seed-0 models; 3-s segments centred on the local artifact where present). (a) MA, (b) EM, (c) BW, and (d) combined BW+MA+EM contamination. CLEAR tracks the clean reference closely in the BW-containing panels (c) and (d), whereas the U-Net output remains smoother under local-only artifacts.}
  \label{fig:qualitative_waveforms}
\end{figure*}

Qualitative inspection of representative fixed-test windows agrees with this contrast (Fig.~\ref{fig:qualitative_waveforms}).

\begin{figure*}[!t]
  \centering
  \includegraphics[width=0.76\textwidth]{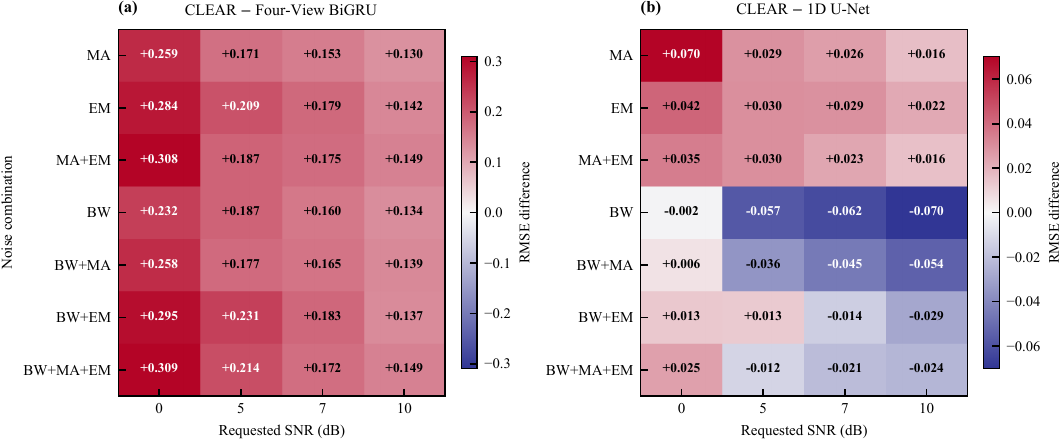}
  \caption{Signed RMSE differences across seven noise combinations and four requested SNR levels. (a) CLEAR minus Four-View BiGRU and (b) CLEAR minus 1D U-Net using common seeds 0 and 1. Negative values (blue) indicate lower RMSE from CLEAR; positive values (red) indicate lower RMSE from the comparator. The two panels use independent color scales matched to their own value ranges. All 28 cell values are annotated directly.}
  \label{fig:noise_cell_rmse}
\end{figure*}

Fig.~\ref{fig:noise_cell_rmse} localizes the comparison across 28 noise--SNR cells. Four-View BiGRU had lower RMSE in every cell. CLEAR had lower two-seed mean RMSE than U-Net in 12 cells and lower RMSE in both seeds in 11: all four BW-only conditions plus eight BW-containing mixtures. At BW-only 10~dB, CLEAR and U-Net obtained RMSE 0.2313 and 0.3014, respectively; at BW--MA--EM 10~dB, values were 0.2875 and 0.3110. U-Net remained better for every local-only MA, EM, or MA--EM cell. CLEAR's advantage is therefore concentrated in baseline-wander conditions, particularly at moderate-to-high requested SNR; these counts are descriptive, not independent significance tests.

\subsection{Morphology and Rhythm Preservation}\label{sec:morphology_results}

Waveform shape and pointwise error are separated. CLEAR obtained QRS-proxy slope error $0.0537\pm0.0012$, derivative correlation $0.9276\pm0.0013$, STFT distance $0.0049\pm0.0000$, matched R-peak timing error $9.7203\pm0.2578$~ms, and median absolute RR error $32.1633\pm1.3788$~ms. The LF/HF difference was highly variable ($3.3971\pm1.5026$), so reliable frequency-domain HRV recovery is not claimed for 10-s windows.

Across NORM, MI, STTC, CD, and HYP, derivative correlation ranged from 0.9125 (CD) to 0.9306 (NORM), STFT distance from 0.0047 to 0.0052, R-peak timing error from 9.043 to 10.206~ms, and slope error from 0.0503 to 0.0557. Morphology consistency therefore did not rest on the largest class alone, although the HYP subset was small ($n=187$) and these are reference-consistency rather than expert clinical measurements.

CLEAR improved R-peak F1 in all six detector--tolerance settings (Table~\ref{tab:rpeak_sensitivity}), with gains from 0.0225 to 0.0593. NeuroKit2 at 150~ms improved from 0.9411 to $0.9636\pm0.0006$; missed and wrong-peak rates decreased from 0.0560 to 0.0288 and from 0.0578 to 0.0408. At 50~ms, F1 still improved from 0.9095 to 0.9333. Pan--Tompkins at 150~ms improved from 0.8669 to 0.9262. These consistent directions strengthen the rhythm result beyond one operating point, but references remain detector-derived and aggregate BiGRU and U-Net F1 remained higher.

Noise-specific analysis retained exceptions to the aggregate gain. MA-only F1 increased from 0.8987 to $0.9529\pm0.0017$ and mean RR error decreased from 50.20 to 35.52~ms; MA--EM and EM-only also improved F1. Conversely, BW-only F1 decreased slightly from 0.9716 to $0.9698\pm0.0007$, and small RR-error increases occurred for BW--MA and EM-only. BW-related RMSE advantages therefore do not imply uniform rhythm improvement.

To remove detector-derived references, the seed-0 model was also scored against the reference annotations distributed with the MIT-BIH Arrhythmia Database, using records 105, 108, 203, and 207, which carry substantial motion artifact. No weight was fine-tuned, and each record was scaled by statistics of its own observed signal, which are available at deployment. Across 180 windows per record, mean R-peak F1 at a 150-ms tolerance increased from $0.796$ to $0.836$. The gain was largest for the most contaminated record, 108, where F1 rose from $0.463$ to $0.561$ ($+9.7$ points) and recall from $0.425$ to $0.552$; the remaining three records improved by $0.4$ to $3.2$ points. PhysioNet distributes these annotations as the \texttt{.atr} reference set; the \texttt{bxb} variant from the original tapes is not publicly available.

\subsection{Localization, Quality, and Selective Inference}\label{sec:auxiliary_results}

Table~\ref{tab:main_auxiliary} evaluates CLEAR's additional outputs. Pointwise mask F1 was $0.8573\pm0.0056$, whereas stricter event-level F1 was $0.3647\pm0.0339$; sample-level inspection does not establish precise clinical event boundaries. Context macro AUROC was $0.9773\pm0.0017$, and global-artifact macro F1 and AUROC were $0.9610\pm0.0125$ and $0.9987\pm0.0008$. Power-line F1 reached $0.9417\pm0.0260$, but recognition must not be confused with the correction failure in Section~\ref{sec:stress_results}. Simulation-derived quality macro F1 was moderate at $0.4768\pm0.0149$; ECE $0.0356\pm0.0022$ does not compensate for quality-state errors.

Quality-guided inference retained $2,554.0\pm48.8$ windows (Table~\ref{tab:main_selective}), giving coverage $0.7702\pm0.0147$ and discard rate $0.2298\pm0.0147$ . On retained windows, adaptive RMSE was $0.3395\pm0.0007$ versus $0.8408\pm0.0047$ when left noisy, and R-peak F1 was 0.9700 versus 0.9543. Patient-level bootstrap differences for seeds 0--2 were $-0.4991$, $-0.5048$, and $-0.5084$, with respective 95\% intervals $[-0.5104,-0.4878]$, $[-0.5164,-0.4945]$, and $[-0.5198,-0.4968]$. These results support a robust retained-patient reduction, not a full-cohort benefit after discarded windows are removed.

All-denoised RMSE on the same retained subset was 0.3372, slightly below adaptive RMSE; the policy's value is explicit pass/discard control rather than extra denoising accuracy. Clean controls gave pass rate $0.9998\pm0.0002$ and alteration RMSE $0.0358\pm0.0039$, but maximum absolute sample change reached $4.1145\pm0.7880$. High pass rates therefore do not guarantee absence of local alteration.

The learned quality head was also compared with a conventional signal-quality baseline. Eight indices were computed per window (kurtosis, skewness, zero-crossing rate, spectral entropy, and four band-power ratios) and classified by logistic regression with balanced class weights under the same leave-one-subject-out protocol. The index baseline reached macro F1 $0.448\pm0.187$ against $0.560\pm0.225$ for the 291-parameter head, an advantage of about eleven points for the learned representation. A label-shuffling probe repeated 100 times reached a selection score of $0.295\pm0.013$, against $0.628$ with the true labels ($p<0.01$), so the signal came from the learned features rather than from the class prior. Class-wise recall was 0.797 for good, 0.806 for noisy, and 0.171 for discard, the last reflecting the 2.8\% discard prior in the pooled folds against a 30.8\% share in the frozen test set.

Whether the masks localize what the model removes was checked directly. The residual energy falling outside the predicted mask was $1.14\%$ of the total residual energy, and the energy outside the true mask was $0.81\%$, so the intervention follows the predicted region rather than a global subtraction. Under a looser event criterion that accepts any predicted event overlapping a reference event with an interval overlap of at least 0.3, precision was $0.677$ and recall $0.893$ (F1 $0.706$), against $0.365$ under the stricter one-to-one matching of Table~\ref{tab:main_auxiliary}; 8{,}774 events were predicted against 3{,}727 reference events, which indicates over-segmentation rather than missed detection.

\subsection{Cross-Database Transfer and Human-Quality Adaptation}\label{sec:transfer_results}

Zero-fine-tuning denoising on identical controlled views (Table~\ref{tab:external_common_seed}). Four-View BiGRU retained the best RMSE, $\Delta$SNR, and R-peak F1 on every external database. CLEAR remained substantially better than the validation-selected traditional filters: its common-seed RMSE was 0.4343, 0.4327, and 0.4309 on QTDB, MITDB, and ECG-ID, compared with bandpass values 0.7810, 0.7133, and 0.9029. Three-seed external R-peak F1 was 0.9564, 0.9616, and 0.9852.

CLEAR reduced RMSE relative to U-Net by 3.26\% on QTDB and 5.64\% on MITDB, but U-Net reduced RMSE by 0.03027 on ECG-ID. CLEAR had lower mean RMSE in 17/28 QTDB cells, 18/28 MITDB cells, and 6/28 ECG-ID cells; Four-View BiGRU was lower in all 84 external cells and all 112 cells including PTB-XL. Transfer is therefore useful but database-dependent, not uniformly superior to compact convolutional denoising.

Auxiliary outputs also transferred with degradation: pointwise mask F1 was 0.7402--0.7541, event F1 0.1708--0.2075, context macro AUROC 0.8860--0.9341, and quality macro F1 0.4694--0.4901. Coverage was 0.7207--0.7366, retained adaptive RMSE 0.3766--0.3795, and clean pass rate 0.9951--1.0000. One forward pass thus retains inspection and quality information across databases without suggesting distribution independence.

BUT QDB tested a separate question: whether quality can be adapted to human annotations without changing denoising. Only $3\times96+3=291$ quality-head parameters (1.03\% of CLEAR) were trained. The split comprised 18,572 training windows from nine subjects, 278 validation windows from one, and 14,109 test windows from five; test annotations were not used for training or selection.

Window-level macro F1 increased from $0.2444\pm0.0310$ to $0.8176\pm0.0178$ (Table~\ref{tab:quality_adaptation}), subject-macro F1 from 0.3513 to 0.6390, and consensus $\kappa$ from 0.0742 to 0.7762, while ECE decreased from 0.4197 to 0.0447. Discard recall increased from 0.0005 to 0.9950 and good recall remained 0.9866. However, noisy recall remained only $0.4405\pm0.0489$, and the five-subject test scope precludes a clinical-grade claim. The result supports modular quality supervision with unchanged denoising weights, not cross-database denoising fine-tuning.

\input{tables/quality_adaptation}

\subsection{Computational Efficiency}\label{sec:efficiency_results}

Resource tradeoffs are as follows (Table~\ref{tab:efficiency}). CLEAR requires 27{,}136 multiply--accumulate operations per 10-s sample (9.77~MMAC/s at 360~Hz), its peak activation footprint is 675~KiB for a full window and 4.69~KiB at a 25-sample chunk, and dynamic INT8 covers 12{,}682 of the 28{,}330 parameters (44.8\%). CLEAR uses 28.23\% of Four-View BiGRU's parameters and 29.32\% of its serialized FP32 storage. On the recorded x86 backend, mean FP32 latency was 61.12~ms versus 451.64~ms, a 7.39-fold advantage that cannot be extrapolated to other processors. CLEAR used 5.34\% fewer parameters and 12.37\% less storage than U-Net, but U-Net was faster (52.03 versus 61.12~ms).

Dynamic INT8 reduced CLEAR's serialized size by 29.08\% to 82.65~KiB without reducing mean latency. Four-View BiGRU showed an extreme backend-specific INT8 slowdown, while U-Net's mean INT8 latency was 49.27~ms. These measurements support compact storage and causal processing, not universal real-time performance or wearable energy efficiency.

%% file: tables/main_morphology.tex
\begin{table}[!t]
  \centering
  \caption{CLEAR Morphology and Rhythm Consistency on the Main Test Set (Seeds 0, 1, and 2; Mean $\pm$ Sample SD). These Are Signal-Consistency Measures, Not Expert Clinical Interval Measurements.}
  \label{tab:main_morphology}
  \scriptsize
  \setlength{\tabcolsep}{2.6pt}
  \renewcommand{\arraystretch}{1.08}
  \begin{tabular}{@{}llr@{}}
    \toprule
    Metric & Unit & CLEAR \\
    \midrule
    PRD & \% & 22.5752 $\pm$ 0.0478 \\
    QRS-proxy slope error & -- & 0.0537 $\pm$ 0.0012 \\
    Derivative correlation & -- & 0.9276 $\pm$ 0.0013 \\
    Log-magnitude STFT distance & -- & 0.0049 $\pm$ 0.0000 \\
    Matched R-peak timing error & ms & 9.7203 $\pm$ 0.2578 \\
    R-peak F1 (150 ms) & -- & 0.9636 $\pm$ 0.0006 \\
    Median absolute RR error & ms & 32.1633 $\pm$ 1.3788 \\
    Absolute SDNN difference & -- & 0.0446 $\pm$ 0.0006 \\
    Absolute RMSSD difference & -- & 0.0623 $\pm$ 0.0009 \\
    Absolute LF/HF difference & -- & 3.3971 $\pm$ 1.5026 \\
    \bottomrule
  \end{tabular}
\end{table}

%% file: tables/main_selective.tex
\begin{table}[!t]
  \centering
  \caption{Selective Inference and Clean Controls (Three Seeds; Mean $\pm$ Sample SD). Retained-Subset Metrics Exclude Discarded Windows.}
  \label{tab:main_selective}
  \scriptsize
  \setlength{\tabcolsep}{2.6pt}
  \renewcommand{\arraystretch}{1.08}
  \begin{tabular}{@{}lr@{}}
    \toprule
    Metric & CLEAR \\
    \midrule
    Retained windows & 2554.0 $\pm$ 48.8 \\
    Coverage & 0.7702 $\pm$ 0.0147 \\
    Discard rate & 0.2298 $\pm$ 0.0147 \\
    Adaptive RMSE on retained & 0.3395 $\pm$ 0.0007 \\
    Noisy RMSE on same retained windows & 0.8408 $\pm$ 0.0047 \\
    All-denoised RMSE on same windows & 0.3372 $\pm$ 0.0011 \\
    Adaptive R-peak F1 on retained & 0.9700 $\pm$ 0.0007 \\
    Noisy R-peak F1 on same windows & 0.9543 $\pm$ 0.0009 \\
    Clean-control RMSE & 0.0358 $\pm$ 0.0039 \\
    Clean-control maximum absolute change & 4.1145 $\pm$ 0.7880 \\
    Clean-control pass rate & 0.9998 $\pm$ 0.0002 \\
    \bottomrule
  \end{tabular}
\end{table}

%% file: tables/efficiency.tex
\begin{table*}[!t]
  \centering
  \caption{Single-Window CPU Efficiency for a $1\times3600\times1$ Input (50 Timed Repetitions After 10 Warm-Ups). Latencies Are Mean $\pm$ Sample SD Across Available Common Seeds. INT8 Uses Dynamic Quantization on the Recorded x86 Backend.}
  \label{tab:efficiency}
  \scriptsize
  \setlength{\tabcolsep}{2.8pt}
  \renewcommand{\arraystretch}{1.08}
  \begin{tabular}{@{}lrrrrrrr@{}}
    \toprule
    & & \multicolumn{4}{c}{FP32} & \multicolumn{2}{c}{INT8} \\
    \cmidrule(lr){3-6} \cmidrule(lr){7-8}
    Method & Params. & Mean (ms) & p50 (ms) & p95 (ms) & KiB & Mean (ms) & KiB \\
    \midrule
    CLEAR & 28,330 & 61.12 $\pm$ 1.07 & 60.99 $\pm$ 0.99 & 61.67 $\pm$ 1.11 & 116.54 & 62.44 $\pm$ 1.12 & 82.65 \\
    Four-View BiGRU & 100,353 & 451.64 $\pm$ 153.67 & 445.92 $\pm$ 153.51 & 486.50 $\pm$ 150.17 & 397.48 & 15226.91 $\pm$ 19226.23 & 110.04 \\
    1D U-Net & 29,928 & 52.03 $\pm$ 2.73 & 69.90 $\pm$ 28.02 & 92.59 $\pm$ 1.53 & 133.00 & 49.27 $\pm$ 4.10 & 133.00 \\
    \bottomrule
  \end{tabular}
\end{table*}

%% file: tables/ablation_main.tex
\begin{table*}[!t]
  \centering
  \caption{Combined Three-Seed Ablation Results. RMSE Entries Are Mean $\pm$ Sample SD across Seeds; R-F1 Denotes R-Peak F1 at 150-ms Tolerance; Context and Global Denote Macro AUROC and Macro F1. Dashes Mark Outputs Whose Supervision Was Removed.}
  \label{tab:ablation_main}
  \scriptsize
  \setlength{\tabcolsep}{1.7pt}
  \renewcommand{\arraystretch}{1.08}
  \begin{tabular}{@{}lrrrrrrr@{}}
    \toprule
    Variant & RMSE & R-F1 & Mask point & Mask event & Context & Global & Clean RMSE \\
    \midrule
    Full CLEAR & 0.3799 $\pm$ 0.0008 & 0.9636 & 0.8573 & 0.3647 & 0.9773 & 0.9610 & 0.0358 \\
    Denoise-only & \textbf{0.3526 $\pm$ 0.0020} & 0.9662 & -- & -- & -- & -- & 0.0478 \\
    w/o mask supervision & 0.3744 $\pm$ 0.0020 & 0.9647 & -- & -- & 0.9683 & 0.9606 & 0.0370 \\
    w/o quality/context & 0.3636 $\pm$ 0.0023 & 0.9654 & 0.8655 & 0.4281 & -- & -- & 0.0377 \\
    TCN-only & 0.5044 $\pm$ 0.0037 & 0.9551 & 0.7488 & 0.0716 & 0.9615 & 0.9216 & 0.0761 \\
    GRU-only & 0.4278 $\pm$ 0.0108 & 0.9591 & 0.8043 & 0.0702 & 0.9317 & 0.6804 & 0.0490 \\
    w/o noise augmentation & 0.4177 $\pm$ 0.0049 & 0.9626 & 0.8382 & 0.3826 & 0.9537 & 0.4108 & 0.0336 \\
    w/o shape regularization & 0.3819 $\pm$ 0.0007 & 0.9633 & 0.8583 & 0.3726 & 0.9776 & 0.9614 & 0.0358 \\
    w/o explanation consistency & 0.3804 $\pm$ 0.0006 & 0.9634 & 0.8582 & 0.3628 & 0.9772 & 0.9612 & 0.0412 \\
    \bottomrule
  \end{tabular}
\end{table*}

%% file: tables/external_common_seed.tex
\begin{table*}[!t]
  \centering
  \caption{Zero-Fine-Tuning External-Database Results. Learned Models Use Common Seeds 0 and 1 (Mean $\pm$ Sample SD); Deterministic References Have No Seed Repetition. All Methods Use Identical Controlled Test Views, No External Denoising Weight Is Updated, and Best Results Are Bold.}
  \label{tab:external_common_seed}
  \scriptsize
  \setlength{\tabcolsep}{3.4pt}
  \renewcommand{\arraystretch}{1.08}
  \begin{tabular}{@{}llrrrr@{}}
    \toprule
    Database & Method & $n$ & RMSE & $\Delta$SNR (dB) & R-peak F1 \\
    \midrule
    QTDB & Noisy input & 1080 & 1.1433 & 0.0000 & 0.9403 \\
     & Wavelet & 1080 & 1.1394 & 0.027 & 0.9384 \\
     & Bandpass & 1080 & 0.7810 & 3.334 & 0.9432 \\
     & CLEAR & 1080 & 0.4343 $\pm$ 0.0011 & 8.084 $\pm$ 0.020 & 0.9564 $\pm$ 0.0027 \\
     & Four-View BiGRU & 1080 & \textbf{0.2129 $\pm$ 0.0073} & \textbf{14.415 $\pm$ 0.277} & \textbf{0.9706 $\pm$ 0.0011} \\
     & 1D U-Net & 1080 & 0.4490 $\pm$ 0.0006 & 7.701 $\pm$ 0.020 & 0.9549 $\pm$ 0.0004 \\
    \midrule
    MITDB & Noisy input & 576 & 1.0842 & 0.0000 & 0.9498 \\
     & Wavelet & 576 & 1.0809 & 0.024 & 0.9484 \\
     & Bandpass & 576 & 0.7133 & 3.681 & 0.9549 \\
     & CLEAR & 576 & 0.4327 $\pm$ 0.0009 & 7.688 $\pm$ 0.007 & 0.9616 $\pm$ 0.0015 \\
     & Four-View BiGRU & 576 & \textbf{0.2238 $\pm$ 0.0060} & \textbf{13.572 $\pm$ 0.202} & \textbf{0.9710 $\pm$ 0.0010} \\
     & 1D U-Net & 576 & 0.4585 $\pm$ 0.0001 & 7.008 $\pm$ 0.004 & 0.9669 $\pm$ 0.0004 \\
    \midrule
    ECG-ID & Noisy input & 1240 & 1.2068 & 0.0000 & 0.9690 \\
     & Wavelet & 1240 & 1.2040 & 0.017 & 0.9685 \\
     & Bandpass & 1240 & 0.9029 & 2.170 & 0.9743 \\
     & CLEAR & 1240 & 0.4309 $\pm$ 0.0037 & 8.595 $\pm$ 0.081 & 0.9852 $\pm$ 0.0002 \\
     & Four-View BiGRU & 1240 & \textbf{0.2293 $\pm$ 0.0055} & \textbf{14.293 $\pm$ 0.170} & \textbf{0.9903 $\pm$ 0.0008} \\
     & 1D U-Net & 1240 & 0.4007 $\pm$ 0.0006 & 9.265 $\pm$ 0.006 & 0.9853 $\pm$ 0.0001 \\
    \bottomrule
  \end{tabular}
\end{table*}

%% file: tables/rpeak_sensitivity.tex
\begin{table}[!t]
  \centering
  \caption{R-Peak Sensitivity on 3,316 Windows. CLEAR F1 and $\Delta$F1 Are Three-Seed Mean $\pm$ Sample SD.}
  \label{tab:rpeak_sensitivity}
  \scriptsize
  \setlength{\tabcolsep}{2.4pt}
  \renewcommand{\arraystretch}{1.08}
  \begin{tabular}{@{}lrrrr@{}}
    \toprule
    Detector & Tol. (ms) & Noisy F1 & CLEAR F1 & $\Delta$F1 \\
    \midrule
    NeuroKit2 & 50 & 0.9095 & 0.9333 $\pm$ 0.0011 & 0.0238 \\
    NeuroKit2 & 100 & 0.9209 & 0.9440 $\pm$ 0.0011 & 0.0231 \\
    NeuroKit2 & 150 & 0.9411 & 0.9636 $\pm$ 0.0006 & 0.0225 \\
    Pan--Tompkins & 50 & 0.6959 & 0.7309 $\pm$ 0.0080 & 0.0350 \\
    Pan--Tompkins & 100 & 0.8032 & 0.8579 $\pm$ 0.0046 & 0.0547 \\
    Pan--Tompkins & 150 & 0.8669 & 0.9262 $\pm$ 0.0022 & 0.0593 \\
    \bottomrule
  \end{tabular}
\end{table}

%% file: tables/main_auxiliary.tex
\begin{table}[!t]
  \centering
  \caption{CLEAR Auxiliary Outputs on the Main Test Set (Three Seeds; Mean $\pm$ Sample SD). Pointwise and Event Mask Scores Are Distinct Evaluations.}
  \label{tab:main_auxiliary}
  \scriptsize
  \setlength{\tabcolsep}{2.6pt}
  \renewcommand{\arraystretch}{1.08}
  \begin{tabular}{@{}lr@{}}
    \toprule
    Output metric & CLEAR \\
    \midrule
    Pointwise mask F1 & 0.8573 $\pm$ 0.0056 \\
    Pointwise mask precision & 0.8793 $\pm$ 0.0044 \\
    Pointwise mask recall & 0.8588 $\pm$ 0.0047 \\
    Event-level mask F1 & 0.3647 $\pm$ 0.0339 \\
    \cmidrule(lr){1-2}
    Three-class quality macro F1 & 0.4768 $\pm$ 0.0149 \\
    Quality ECE & 0.0356 $\pm$ 0.0022 \\
    Noise-context macro AUROC & 0.9773 $\pm$ 0.0017 \\
    Global-artifact macro F1 & 0.9610 $\pm$ 0.0125 \\
    Global-artifact macro AUROC & 0.9987 $\pm$ 0.0008 \\
    Global BW F1 & 0.9804 $\pm$ 0.0023 \\
    Power-line F1 & 0.9417 $\pm$ 0.0260 \\
    \bottomrule
  \end{tabular}
\end{table}

%% file: tables/main_denoising.tex
\begin{table}[!t]
  \centering
  \caption{Main Denoising Performance on the Fixed PTB-XL Test Set (3,316 Windows). Learned Models Use Common Seeds 0 and 1; Values Are Mean $\pm$ Sample SD. Deterministic References Have No Seed Repetition. Single-View and Four-View BiGRU Share 100,353 Parameters; Four-View Training Changes the Data Protocol, Not the Network. Best Results Are Bold.}
  \label{tab:main_denoising}
  \scriptsize
  \setlength{\tabcolsep}{1.8pt}
  \renewcommand{\arraystretch}{1.08}
  \begin{tabular}{@{}lrrr@{}}
    \toprule
    Method & RMSE & $\Delta$SNR (dB) & R-peak F1 \\
    \midrule
    Noisy input & 1.0531 & 0.0000 & 0.9411 \\
    Wavelet & 1.0493 & 0.0302 & 0.9398 \\
    Bandpass & 0.7040 & 3.5039 & 0.9457 \\
    \midrule
    Single-View BiGRU & 0.3773 $\pm$ 0.0032 & 9.7017 $\pm$ 0.0676 & 0.9477 $\pm$ 0.0023 \\
    Dias official GRU & 1.1752 $\pm$ 0.0010 & -1.7057 $\pm$ 0.0084 & 0.7472 $\pm$ 0.0084 \\
    Four-View BiGRU & \textbf{0.1863 $\pm$ 0.0065} & \textbf{14.8006 $\pm$ 0.2954} & \textbf{0.9788 $\pm$ 0.0004} \\
    1D U-Net & 0.3619 $\pm$ 0.0006 & 8.8089 $\pm$ 0.0180 & 0.9689 $\pm$ 0.0001 \\
    CLEAR (ours) & 0.3795 $\pm$ 0.0002 & 8.4759 $\pm$ 0.0029 & 0.9636 $\pm$ 0.0009 \\
    \bottomrule
  \end{tabular}
\end{table}

%% file: tables/quality_adaptation.tex
\begin{table}[!t]
  \centering
  \caption{Independent BUT QDB Quality-Head Adaptation on 14,109 Test Windows From Five Subjects (Three Seeds; Mean $\pm$ Sample SD). Only the Three-Class Quality Head Is Adapted; Denoising Weights Are Unchanged.}
  \label{tab:quality_adaptation}
  \scriptsize
  \setlength{\tabcolsep}{2.2pt}
  \renewcommand{\arraystretch}{1.08}
  \begin{tabular}{@{}lrr@{}}
    \toprule
    Metric & Before & After \\
    \midrule
    Window-level quality macro F1 & 0.2444 $\pm$ 0.0310 & 0.8176 $\pm$ 0.0178 \\
    Subject-macro quality F1 & 0.3513 $\pm$ 0.0174 & 0.6390 $\pm$ 0.0679 \\
    Quality ECE & 0.4197 $\pm$ 0.0128 & 0.0447 $\pm$ 0.0086 \\
    Consensus $\kappa$ & 0.0742 $\pm$ 0.0477 & 0.7762 $\pm$ 0.0168 \\
    Good-class recall & 1.0000 $\pm$ 0.0000 & 0.9866 $\pm$ 0.0047 \\
    Noisy-class recall & 0.0048 $\pm$ 0.0035 & 0.4405 $\pm$ 0.0489 \\
    Discard-class recall & 0.0005 $\pm$ 0.0001 & 0.9950 $\pm$ 0.0012 \\
    Mean annotator-agreement $\kappa$ & 0.0862 $\pm$ 0.0546 & 0.7800 $\pm$ 0.0134 \\
    \bottomrule
  \end{tabular}
\end{table}

%% file: sections/05_ablation_discussion.tex
\Needspace{8\baselineskip}
\section{Ablation, Stress Testing, and Discussion}\label{sec:ablation_discussion}

\subsection{Ablation Studies}\label{sec:ablation_results}

\begin{figure}[!t]
  \centering
  \includegraphics[width=0.98\columnwidth]{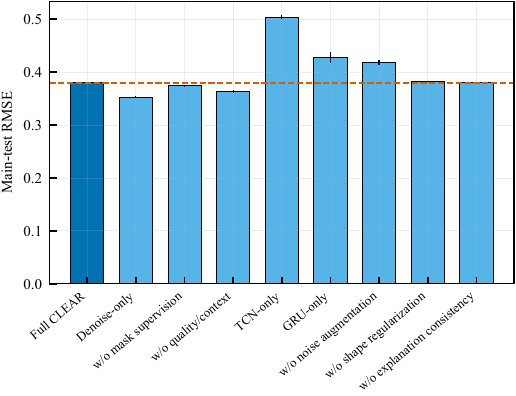}
  \caption{Three-seed ablation results: main-test RMSE with sample-SD error bars. The blue bar and dashed line denote Full CLEAR. Pointwise and event-level mask F1 for all variants are reported with the principal comparisons in Table~\ref{tab:ablation_main}; denoise-only and w/o-mask variants have no trained mask outputs.}
  \label{fig:ablation_results}
\end{figure}

Variants were compared using identical sample identifiers and seed-specific counterparts. Paired $\Delta$RMSE is variant minus Full CLEAR for the same seed, averaged across three seeds with its own sample SD across seeds; negative values favor the variant, and this SD differs from the per-variant RMSE SD given in the table. The principal comparisons are summarized below (Table~\ref{tab:ablation_main}), while Fig.~\ref{fig:ablation_results} highlights the principal comparisons.

Denoise-only produced the lowest ablation RMSE ($0.3526\pm0.0020$), a paired improvement of $-0.0273$ or 7.19\% over Full CLEAR, and increased $\Delta$SNR to 9.104~dB and R-peak F1 to 0.9662. Removing mask supervision improved RMSE by 0.0055; removing quality/context supervision improved it by 0.0163 and increased pointwise/event mask F1 to 0.8655/0.4281. Auxiliary supervision therefore imposes a measurable reconstruction tradeoff rather than uniformly improving every task. Denoise-only has no trained mask or quality function, and removing quality/context eliminates the proposed decision interface; their untrained outputs are not scored.

Both backbone stages were necessary for the full operating profile. TCN-only increased RMSE by $0.1244\pm0.0034$ to 0.5044 and reduced event mask F1 to 0.0716; GRU-only increased RMSE by $0.0479\pm0.0101$ to 0.4278 and reduced global-artifact macro F1 to 0.6804. Their clean-control RMSE also increased to 0.0761 and 0.0490. Restoring the complete model reduced RMSE by 24.67\% relative to TCN-only and 11.19\% relative to GRU-only.

Removing augmentation increased RMSE by $0.0377\pm0.0056$ to 0.4177, reduced context AUROC from 0.9773 to 0.9537, and reduced global-artifact macro F1 from 0.9610 to 0.4108. External RMSE increased on QTDB (0.4355 to 0.4841), MITDB (0.4330 to 0.4642), and ECG-ID (0.4324 to 0.4808). Coverage rose to 0.9142, but higher retention is not inherently preferable when artifact recognition deteriorates. Shape regularization and explanation consistency changed paired RMSE by only $+0.0019$ and $+0.0005$; three seeds establish no substantial independent contribution from either.

The external figures quoted in this subsection are three-seed means, whereas Table~\ref{tab:external_common_seed} lists the common-seed values used for cross-method comparison. The external results preserved the same tradeoff. Denoise-only RMSE was 0.3986, 0.4009, and 0.3965 on QTDB, MITDB, and ECG-ID, below Full CLEAR but without trained localization or quality outputs. TCN-only, GRU-only, and no-augmentation variants generally degraded external RMSE and mask performance. Full CLEAR is thus not the reconstruction-optimal member of its ablation family; its distinction is competitive causal denoising together with inspectable and quality-related outputs.

\subsection{Stress Tests and Failure Cases}\label{sec:stress_results}

Under time-varying SNR, CLEAR's RMSE was $0.3881\pm0.0005$ and $D_k=1.025\pm0.001$, close to its base-condition output; Four-View BiGRU and U-Net had ratios 1.017 and 1.012. Across burst, packet loss, electrode pop, saturation, and power-line interventions, $D_k<1$ denotes lower output error than the original base-input output, not improvement over the newly stressed input. Four-View BiGRU retained the lowest stressed RMSE in every condition, and U-Net was lower than CLEAR in most. These tests characterize sensitivity rather than establish a robustness ranking in CLEAR's favor.

The dedicated power-line experiment exposes a clear performance boundary. Table~\ref{tab:powerline_failure} shows that the contaminated input already had low RMSE (0.0413), whereas CLEAR increased it to $0.0936\pm0.0035$, with $\Delta$SNR $-6.660\pm0.376$~dB. R-peak F1 decreased slightly at every tolerance, including 0.9805 to $0.9793\pm0.0010$ at 150~ms. CLEAR therefore worsened this weak, predominantly power-line-contaminated condition despite high power-line classification performance.

\begin{figure}[!t]
  \centering
  \includegraphics[width=0.98\columnwidth]{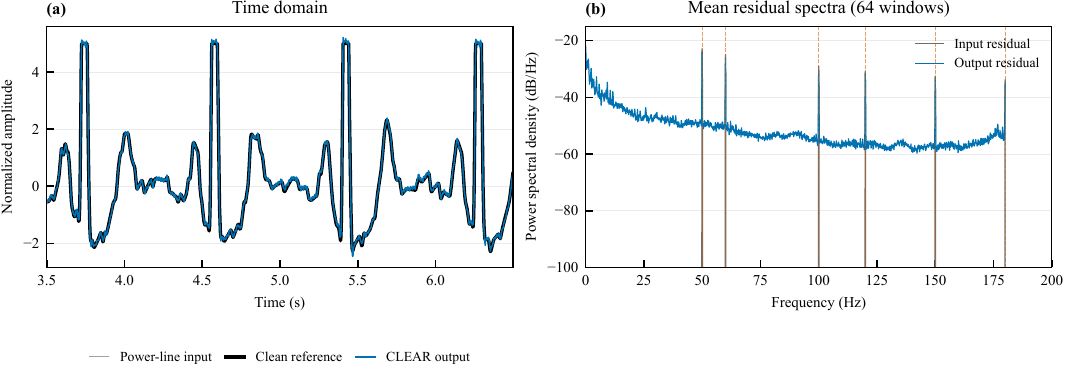}
  \caption{Power-line failure case. (a) A 3-s segment of a power-line-contaminated fixed-test window (seed 0): the contaminated input is already close to the clean reference, and CLEAR's output deviates from it. (b) Mean Hann-periodogram spectra of input and output residuals across 64 stressed windows. Narrowband suppression is visible at the 50/60-Hz fundamentals and harmonics (dashed markers), but the broadband output-residual floor rises above the input floor, dominating the negative overall $\Delta$SNR.}
  \label{fig:powerline_failure}
\end{figure}

Residual power was nevertheless suppressed at targeted frequencies (Fig.~\ref{fig:powerline_failure}). Mean suppression across the six fundamental/harmonic conditions ranged from 1.170~dB at 100~Hz to 3.460~dB at 180~Hz, using Hann-periodogram integration over $\pm0.5$~Hz. Positive narrowband suppression alongside negative overall $\Delta$SNR indicates that off-band or other waveform distortion dominates. Power-line detection remains useful for auditing, but the current model requires a conservative fallback or separately validated correction pathway.

\input{tables/powerline_failure}

A gated fallback was evaluated on the same power-line stress set. When a spectral detector flags line interference, the input is routed through second-order IIR notch filters at the fundamental and its first two harmonics instead of being reconstructed. The detector fired on $69.1\%$ of the 256 stressed windows. Where it fired, the notch path restored $\Delta$SNR to $+1.27$~dB and RMSE to $0.040$, close to the uncontaminated input ($0.041$), whereas CLEAR alone on this subset reached $-6.49$~dB and $0.092$, consistent with the three-seed result in Table~\ref{tab:powerline_failure}; R-peak F1 changed from $0.979$ to $0.982$. The fallback is a conservative switch rather than a substitute for CLEAR, and its detection rate is reported as a limitation of the simple detector.

The error modes observed on the fixed test are consistent across the experiments. Mask predictions over-segment: 8{,}774 events were proposed against 3{,}727 reference events, so precision limits event-level F1 (0.706) while recall remains high (0.893). Discard-class recall was only 0.171 under the 33.8\% test prior, because the head was trained where discard accounts for 2.8\% of windows; the operating point, not the representation, causes this asymmetry. Rhythm regressions were confined to the baseline-wander condition, where F1 fell slightly from 0.9716 to 0.9698 while MA and MA--EM improved. Power-line interference remained the clearest failure, and external auxiliary outputs degraded by roughly a factor of two in event F1 (0.17--0.21) although pointwise mask F1 stayed near 0.75. These modes delimit when the outputs should be trusted.

\subsection{Matched Causal Baseline and Training-Schedule Ablation}\label{sec:single_stage}

Table~\ref{tab:controls} collects the controlled comparisons reported in this section. Two model controls ask whether the observed behaviour follows from the causal convolution front end or from the training schedule; both used the identical four-view data, objective, and checkpoint-selection rule. A third check verifies that chunked inference reproduces the full-window output.

\input{tables/control_experiments}

\subsubsection{Parameter-matched unidirectional GRU}

The first control replaces the four dilated convolutions with a single unidirectional GRU of width 93, giving 28,468 parameters against 28,330 for CLEAR (a $0.5\%$ difference). On the fixed test it reached RMSE $0.3754$ and $\Delta$SNR $8.585$~dB, a $1.2\%$ lower reconstruction error than CLEAR's $0.3799$ and $8.461$~dB. The difference appeared in the auxiliary outputs: R-peak F1 at 50~ms $0.8932$ versus $0.9333$, mask F1 $0.8368$ versus $0.8573$, quality macro F1 $0.4598$ versus $0.4768$, and global-artifact F1 $0.4852$ versus $0.9610$. At an equal parameter budget, therefore, waveform reconstruction is largely insensitive to the front end, and whole-window auditing did not follow from recurrence alone in this comparison. The deficit is not uniform: baseline-wander F1 was $0.9704$ against $0.9804$, whereas power-line F1 was $0.0000$ against $0.9417$ (AUROC $0.6695$ versus $0.9987$), so the recurrent model learned narrowband interference only weakly and never crossed the decision threshold. Because it was retrained with a conservative learning rate and tighter clipping after an initial divergence, and is a single run, the architectural statement is restricted to this comparison.

\subsubsection{Single-stage training at the same budget}

The second control removes the pretraining stage, starting from random initialization and training the full objective for 180 epochs. It is a budget-matched control rather than an isolation of pretraining: the run had not converged at the epoch limit, so the comparison cannot separate the absence of pretraining from a shorter effective optimisation. It reached RMSE $0.4320$, $\Delta$SNR $7.406$~dB, R-peak F1 at 50~ms $0.9267$, and mask F1 $0.8310$, all below the two-stage schedule. Validation RMSE was still decreasing at epoch 180, and quality macro F1 was in fact higher ($0.4946$ versus $0.4768$), so the outcome is mixed rather than uniformly worse. We therefore report this control as an optimisation-budget comparison and draw no conclusion about the necessity of pretraining.

\subsubsection{Chunked inference equivalence}

Streaming deployment requires that chunked inference reproduce the full-window forward pass. A stateful implementation was built that carries the tail buffer of each convolution and the hidden state of the recurrent layer across chunks and was compared with the full-window path on 16 fixed-test windows at chunk lengths of 25, 50, 100, 200, 900, and 3,600 samples. The maximum absolute difference of the reconstructed waveform was $5.84\times10^{-6}$ for chunk lengths up to 200 samples and $4.77\times10^{-6}$ at 900 samples, both below the $10^{-5}$ float32 criterion. The mask and quality logits differed by $1.6\times10^{-5}$ and $7.2\times10^{-6}$ respectively, which is the rounding difference between per-chunk and single-pass accumulation.

\subsection{Discussion}\label{sec:discussion}

The results distinguish reconstruction accuracy from operational value. Four-view training improved the unchanged 100,353-parameter BiGRU by 50.62\%, confirming the contamination protocol. Its remaining advantage may reflect capacity, bidirectional context, and reconstruction-only training; these factors are not isolated. U-Net also exceeded aggregate CLEAR reconstruction at a similar budget, whereas CLEAR was stronger in BW-containing cells. Four-View BiGRU is thus the best offline reconstructor and U-Net the stronger compact aggregate baseline. CLEAR's distinction is causal processing integrated with residual/mask inspection, noise context, global-artifact detection, and quality decisions, rather than uniformly lower error.

The ablations quantify this integration cost. Denoise-only improves RMSE but removes audit functions. Both convolution and recurrence support reconstruction and auxiliary outputs; regularization effects are too small for strong independent claims. Masks, residuals, context, and global outputs explain model intervention, but weak event boundaries and moderate simulation-quality performance limit interpretation.

Selective inference is a coverage decision, not an accuracy shortcut. It retained approximately 77.0\% of main-test windows and improved them relative to noisy input, although all-denoised retained windows were slightly better. Near-unity clean pass rates support bypassing processing, but rare large changes justify preserving the originals for manual review. BUT QDB confirms independent adaptation of 291 quality-head parameters; five test subjects and noisy recall of 0.4405 limit the claim.

External transfer is useful but database-dependent: CLEAR outperformed U-Net on QTDB and MITDB but not ECG-ID, while Four-View BiGRU remained strongest. Augmentation ablation degraded main and external performance without isolating individual perturbation effects. Power-line recognition and harmonic suppression coexisted with worse waveforms, requiring a conservative or separately validated pathway.

\subsection{Limitations}\label{sec:limitations}

\noindent\textbf{Controlled contamination.} The principal benchmark is simulation-based. Measured MA, EM, and BW are combined at controlled locations and amplitudes with device disturbances. Patient/noise isolation prevents the evaluated leakage routes but cannot reproduce every electrode interface, movement, electronic component, or clinical environment. External databases use controlled contaminated views, demonstrating transfer to new clean-source populations rather than prospective validation on real wearable artifacts.

\noindent\textbf{Window and reference scope.} CLEAR uses single-lead 10-s windows at 360~Hz after robust scaling and offline reference preparation. Convolution and recurrence exclude future samples, but pooling and quality decisions require complete windows. The claim is causal sequence processing with segment-level decisions, not sample streaming. Clean references are processed waveforms, R-peak truth is detector-derived, and morphology metrics are not expert annotations. Multilead inputs, variable durations, other sampling rates, and acquisition latency require evaluation.

\noindent\textbf{Quality and selective inference.} Simulation quality labels differ from human judgments; BUT QDB adaptation covers five test subjects with weak noisy-class recall. Diagnostic-classification and patient-outcome benefits were not evaluated. Selective inference discarded approximately 23.0\% of main-test windows, so retained metrics and bootstrap intervals describe neither discarded windows nor clinical missing-data costs.

\noindent\textbf{Comparative and hardware evidence.} Most stressors are synthetic and separately applied; degradation ratios do not measure improvement over stressed inputs. Four-View BiGRU differs in capacity, bidirectionality, outputs, and objective. Two-seed comparators and three-seed CLEAR summaries limit statistical power. Small regularization effects and the two-stage schedule lack dedicated significance or initialization comparisons. Computational evidence is specific to the recorded x86 implementation. Target-device memory traffic, energy, buffer management, and clinically important episodes remain unevaluated.

%% file: tables/powerline_failure.tex
\begin{table}[!t]
  \centering
  \caption{Dedicated Power-Line Stress Test (Three Seeds; Mean $\pm$ Sample SD). R-Peak Matching Uses the Primary 150-ms Tolerance; the 50-ms and 100-ms Settings Give the Same Direction and Are Summarized in the Text. Negative $\Delta$SNR and $\Delta$F1 Are Retained as Explicit Failure Results.}
  \label{tab:powerline_failure}
  \scriptsize
  \setlength{\tabcolsep}{1.6pt}
  \renewcommand{\arraystretch}{1.08}
  \begin{tabular}{@{}lrrrr@{}}
    \toprule
    Input RMSE & CLEAR RMSE & $\Delta$SNR (dB) & Noisy F1 & CLEAR F1 \\
    \midrule
    0.0413 & 0.0936 $\pm$ 0.0035 & -6.660 $\pm$ 0.376 & 0.9805 & 0.9793 $\pm$ 0.0010 \\
    \bottomrule
  \end{tabular}
\end{table}

%% file: tables/control_experiments.tex
\begin{table}[!t]
  \centering
  \caption{Controlled Comparisons on the Fixed Test. R-Peak F1 Uses 50-ms Tolerance; Global F1 Is the Macro Average over the Baseline-Wander and Power-Line Heads. CLEAR Is a Three-Seed Mean with Sample SD 0.0008 (RMSE), 0.03~\textnormal{dB} ($\Delta$SNR), 0.0011 (R-peak), 0.0056 (mask), and 0.0125 (global); Each Control Is a Single Run.}
  \label{tab:controls}
  \scriptsize
  \setlength{\tabcolsep}{2.0pt}
  \renewcommand{\arraystretch}{1.08}
  \begin{tabular}{@{}lrrrrrr@{}}
    \toprule
    Model & Params & RMSE & $\Delta$SNR & R-peak F1 & Mask F1 & Global F1 \\
    \midrule
    CLEAR & 28,330 & 0.3799 & 8.46 & 0.9333 & 0.8573 & 0.9610 \\
    Matched GRU & 28,468 & 0.3754 & 8.585 & 0.8932 & 0.8368 & 0.4852 \\
    Single-stage & 28,330 & 0.4320 & 7.406 & 0.9267 & 0.8310 & 0.9281 \\
    \bottomrule
  \end{tabular}
\end{table}

%% file: sections/06_conclusion.tex
\section{Conclusion and Future Work}\label{sec:conclusion}

This work asked whether a 28,330-parameter causal model can deliver artifact removal that remains auditable under leakage-controlled evaluation. Four conclusions follow from the experiments.

First, reconstruction accuracy at this parameter budget is largely architecture-insensitive, whereas auditing is not: replacing the causal convolution front end with a parameter-matched unidirectional GRU left RMSE within $1.2\%$ but halved global-artifact F1, from $0.961$ to $0.485$. Second, the requirement for sample-level causality does not conflict with offline consistency: chunked inference reproduced the full-window waveform to within $5.8\times10^{-6}$ in float32 while carrying only past context. Third, quality-guided processing was shown to act on the retained subset rather than the full cohort, and adaptation of the 291-parameter quality head to human annotations raised macro F1 from $0.2444$ to $0.8176$ while the denoising weights remained frozen. Fourth, oracle scaling does not change the conclusions: fixed training-split statistics changed the four metrics modestly and in mixed directions; because the RMSE change was about eleven times the three-seed standard deviation, it is reported as a direction rather than as equivalence, and the oracle-scaled model is kept as the main protocol for comparability with earlier results.

The main unresolved questions concern scope rather than accuracy. A diagnostic-band reference has not been established, because a 0.05~Hz high-pass is not identifiable on 10-s windows and would require longer recordings. Human quality annotations remain limited to one database, and a larger cohort with subject-level cross-validation would be needed before the three-state output could be treated as a screening decision. Deployment-side latency, memory, and energy measurements on an embedded target also remain outstanding.

\subsection*{Code and Data Availability}
The data-preparation and evaluation utilities, together with the reproduction instructions, are available at \url{https://github.com/libowen0222/CLEAR-ECG}. The complete implementation, including the model, training configuration, and the supplementary analyses, will be released upon acceptance of this manuscript. Processed splits can be rebuilt from the public PhysioNet records~\cite{moody1984noise,laguna1997qt,moody2001mitdb,wagner2020ptbxl} using the released scripts.